\documentclass[runningheads]{llncs}

\usepackage{eccv}

\usepackage{eccvabbrv}

\usepackage{graphicx}
\usepackage{booktabs}
\usepackage{multirow}
\usepackage{multicol}
\usepackage{amsmath}
\usepackage{amssymb}
\usepackage{cite}
\usepackage{tabu}
\usepackage{tabularx}
\usepackage[utf8]{inputenc}
\usepackage[T1]{fontenc}
\usepackage[dvipsnames]{xcolor}
\usepackage{color, colortbl}
\usepackage[font=small]{caption}
\usepackage{subcaption}
\usepackage[labelformat=simple]{subcaption}

\usepackage{amsmath}
\usepackage{amssymb}
\usepackage[ruled,vlined]{algorithm2e}
\usepackage{booktabs}
\usepackage{tikzscale}
\usepackage{pgfplots}
\usepackage[mode=buildnew]{standalone}
\usepgfplotslibrary{groupplots}
\usepackage{pgfplots, pgfplotstable}
\pgfplotsset{compat=1.17}
\usepackage[accsupp]{axessibility}  
\usepackage{bbm}
\newcolumntype{P}[1]{>{\centering\arraybackslash}p{#1}}

\usepackage{hyperref}

\usepackage{orcidlink}

\newcommand{\putindex}[3]{\vtop{\hbox{\hspace{#3} $#1$}
            \hbox{\raise 6mm \hbox{$\scriptscriptstyle #2$}}}}

\newcommand{\gradx}[0]{\vtop{\hbox{\rm grad}
            \hbox{\raise 2.5mm \hbox{\rm \hspace{2mm} \footnotesize x}}}}

\newcommand{\grady}[0]{\vtop{\hbox{\rm grad}
            \hbox{\raise 2.5mm \hbox{\rm \hspace{2mm} \footnotesize y}}}}

\newcommand{\grad}[1]{\vtop{\hbox{\rm grad}
            \hbox{\raise 2.5mm \hbox{#1}}}}

\newcommand{\stz}{\rule{0mm}{2.3ex}}

\newcommand{\stzdown}{\rule[-1.2ex]{0mm}{3.5ex}}

\newcommand{\btb}{     \begin{tabbing}             }
\newcommand{\bte}{     \end{tabbing}               }

\begin{document}

\title{Foundation Models for Amodal Video Instance Segmentation in Automated Driving} 

\titlerunning{Foundation Models for Amodal VIS in Automated Driving}

\author{Jasmin Breitenstein \inst{1} \and
Franz J\"unger \inst{1} \and Andreas B\"ar \inst{1} \and
Tim Fingscheidt \inst{1}}

\authorrunning{J. Breitenstein et al.}

\institute{$^1$Institute for Communications Technology \\
Technische Universit\"at Braunschweig \\
\email{\{j.breitenstein, f.juenger, andreas.baer, t.fingscheidt\}@tu-bs.de}}

\maketitle

\begin{abstract}
 In this work, we study amodal video instance segmentation for automated driving. Previous works perform amodal video instance segmentation relying on methods trained on entirely labeled video data with techniques borrowed from standard video instance segmentation. 
 Such amodally labeled video data is difficult and expensive to obtain and the resulting methods suffer from a trade-off between instance segmentation and tracking performance. 
To largely solve this issue, we propose to study the application of foundation models for this task. More precisely, we exploit the extensive knowledge of the Segment Anything Model (\texttt{SAM}), while fine-tuning it to the amodal instance segmentation task. Given an initial video instance segmentation, we sample points from the visible masks to prompt our amodal \texttt{SAM}. We use a point memory to store those points. If a previously observed instance is not predicted in a following frame, we retrieve its most recent points from the point memory and use a point tracking method to follow those points to the current frame, together with the corresponding last amodal instance mask. This way, while basing our method on an amodal instance segmentation, we nevertheless obtain video-level amodal instance segmentation results. Our resulting \texttt{S-AModal} method achieves state-of-the-art results in amodal video instance segmentation while resolving the need for amodal video-based labels. Code for \texttt{S-AModal} is available at \url{https://github.com/ifnspaml/S-AModal}. 
 
\end{abstract}

    \vspace{-25pt}
\section{Introduction}
\label{sec:intro}
    \vspace{-2pt}

\begin{figure}[t]
    \centering
    \includegraphics[width=\textwidth]{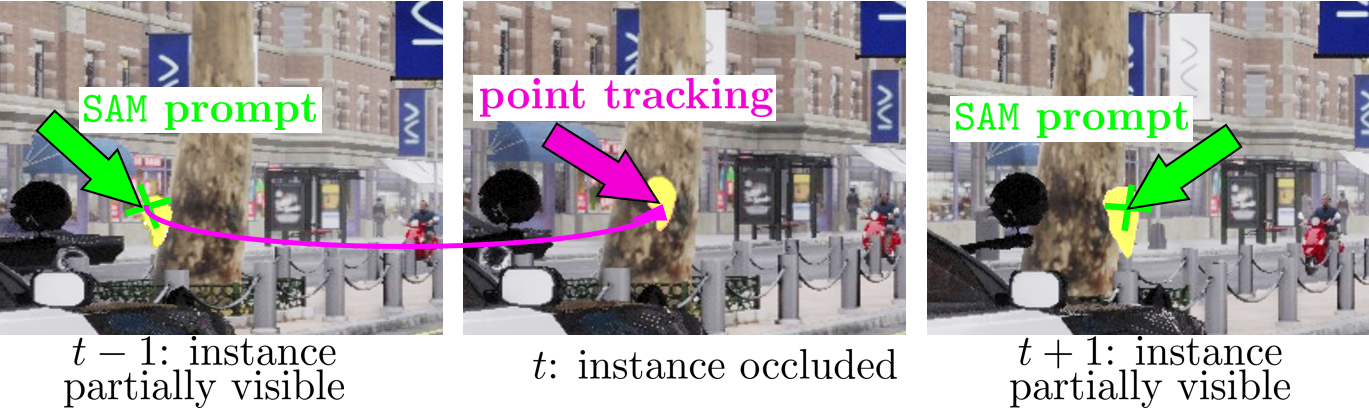}
    \vspace*{-16pt}
    \caption{In a video sequence, when an instance is (partially) visible (time step $t-1$), we extract points (green arrow) from the predicted instance mask to prompt an amodal \texttt{SAM} method to generate an amodal mask (yellow). The corresponding points are stored, and if the instance is not visible (time step $t$), we track the previous points to the next frame ($t$), transferring the previous amodal mask to the next frame (yellow, purple arrow). Once the instance reappears (time step $t+1$), we prompt the amodal \texttt{SAM} method again (green arrow).}
    \label{fig:motivation}
        \vspace{-15pt}
\end{figure}

For safe automated driving, environment perception is required to perform at least on par with human drivers \cite{fingscheidt_dnndataautomateddriving}. 
Recently, foundation models for environment perception have been introduced.
Such models have been trained on a broad amount of data to perform their respective perception task on a wide range of data showcasing impressive zero-shot generalizability.
A famous example for such foundation models is the Segment Anything Model (\texttt{SAM}) \cite{Kirillov2023} which can be prompted via points, bounding boxes, or text, to segment instances in a zero-shot fashion in a wide range of images.
In contrast to this type of perception, one important human ability is amodal perception allowing us to perceive the full shape of occluded objects. 
This ability is typically not found in machine learning perception methods, and, even worse, performance drastically drops when confronted with occlusions.
However, handling occlusions is of high importance for safety in automated driving.
Without a safe treatment of occlusions, one risks fatal accidents of the automated vehicle affecting both vehicle occupants and other traffic participants. 
Amodal segmentation methods offer a way to do this by segmenting not only what is visible, but the full shape of objects in a scene, thus, providing information about what is currently occluded. 
While there exist many methods working towards amodal segmentation on single images --- and excelling at it \cite{Ke2021,Follmann2019,Ling2020,Qi2019,Back2022,Ao2023,Ao2024}, they are limited in the occlusion types they can resolve: Only partial occlusions can be identified and segmented on a single image basis. 
For occlusions that arise temporarily leading to almost no visible object parts or even to a totally occluded object, amodal segmentation methods working on a single image basis cannot rely on sufficient information.
Here, recent new approaches have been introduced and investigated on the task of amodal video instance segmentation (VIS) \cite{Breitenstein2023,Yao2022}. 
Amodal VIS builds upon the terminology of standard (visible) VIS and aims to track and segment the full shape on an instance throughout all frames of a video.
Recent approaches have leveraged a full end-to-end training of amodal VIS \cite{Breitenstein2023}, or investigated a self-supervised training approach based on the results of a VIS method on the considered data \cite{Yao2022}.
In contrast, our proposed \texttt{S-AModal} method relies during training on image-level amodal labels to fine-tune a foundation model to the task of (image-based) amodal segmentation. 
We build upon the highly generalizing pre-trained \texttt{SAM} \cite{Kirillov2023} using an adaptive fine-tuning approach to facilitate the prediction of amodal masks. 
To extend this to videos, during inference a pre-trained VIS method provides point prompts for our amodal \texttt{SAM} method. 
Additionally, we exploit a point tracking method to provide point prompts for occluded instances by relying on a point memory to store points of previously observed instances.
This high-level operation of our \texttt{S-AModal} method is illustrated in Figure \ref{fig:motivation}. 
Whenever an instance is at least partially visible and predicted by the VIS method (yellow, left and right images in Figure \ref{fig:motivation}, time instants $t-1$, $t+1$), we extract points from the visible mask to prompt the \texttt{SAM} method fine-tuned on amodal data (highlighted by a green arrow). In Figure \ref{fig:motivation} one example point prompt is visualized by the green arrow and cross.  
If a previously predicted instance is not predicted for the current frame (purple, middle image of Figure \ref{fig:motivation}), we apply a point tracking method to track the previous ($t-1$) point prompt to the current frame $t$, and we move the amodal mask along the point trajectory.
A reappearing instance is identified by the VIS method, so in this case the \texttt{SAM} method can again be prompted.
Our amodal VIS method based on these foundation models is able to perform state-of-the-art amodal VIS \textit{without} relying on amodal VIS labels in training. 

Our contributions are as follows: 
First, we provide a fine-tuning strategy to leverage \texttt{SAM} for amodal segmentation. 
Second, we show that our proposed \texttt{S-AModal} method achieves state-of-the-art results in image-level amodal segmentation.
Third, we are the first to apply such a foundation model to the task of amodal VIS, again achieving state-of-the-art results in this task.
The remainder of this work is structured as follows: 
In Section \ref{sec:rel-work}, we review the related work w.r.t.\ \texttt{SAM} \cite{Kirillov2023}, point tracking, and amodal segmentation. 
Section \ref{sec:method} describes our proposed method in detail.
In Section \ref{sec:experimental-setup}, we describe the experimental setup and report our results in Section \ref{sec:experiments}.
Finally, we conclude with Section \ref{sec:conclusion}.

    \vspace{-4pt}
\section{Related Work}
\label{sec:rel-work}
    \vspace{-2pt}

Here, we review works related to \texttt{SAM}, point tracking, and amodal segmentation.

\noindent \textbf{Segment Anything Model} (\texttt{SAM}) \cite{Kirillov2023} was first introduced as a foundation model that solves the task of promptable segmentation. 
It can be prompted using text, bounding boxes or points to produce a segmentation mask accordingly. 
\texttt{SAM} has been trained on an extensive amount of data to perform promptable segmentation in a zero-shot manner on many different types of images.

Adapters have been proposed to allow \texttt{SAM} to, e.g., work well on high-quality images \cite{Ke2023,Xie2024}, or, in general, on underperformed scenes \cite{Wu2023,Xiao2024,Chen2023,Shaharabany2023}. 
The adapters follow fine-tuning strategies, so only the decoder and the prompt encoder \cite{Shaharabany2023} or added adapter layers of the encoder \cite{Chen2023} have to be fine-tuned. 
Most related work about fine-tuning \texttt{SAM} is related to underperformed scenes, e.g., in the medical field \cite{Chen2023,Shaharabany2023,Wu2023}.
In contrast, we investigate \texttt{SAM} not only on a different image domain, but also tailor it to the new task of amodal segmentation.

Additionally, strategies have been proposed to extend the \texttt{SAM} capabilities to video sequences \cite{Rajic2023}. 
Rajic et al.~\cite{Rajic2023} perform video segmentation by applying a point tracking method to points of a query mask in the first frame to track it through an entire video sequence, Yang et al.~\cite{Yang2023} use \texttt{SAM} to improve the segmentation of a VIS method.
Cheng et al.~\cite{Cheng2023} extend \texttt{SAM} to track objects via text prompts.
Recently, \texttt{SAM 2} \cite{Ravi2024}, the successor of \texttt{SAM}, has been introduced adding video segmentation capabilities to \texttt{SAM}.
In this work, we build upon the benefits of the above by following a combination of \texttt{SAM} with VIS methods to perform the tracking and segmentation task. 
However, to resolve occlusions, we additionally add a point tracking method to identify points for occluded masks.

\noindent \textbf{Point Tracking} has first been introduced and defined by Doersch et al.~\cite{Doersch2022}, naming it the ``tracking any point'' (TAP) task. This task focuses on predicting the pixel positions of a given point in all subsequent frames of a video sequence, as well as identifying its occlusion state.
Harley et al.\ \cite{Harley2022} develop \texttt{PIPs}, one of the initial methods for the TAP task.
Although \texttt{PIPS} can trace points through temporal occlusions lasting up to eight frames, it cannot account for the occlusion state of the points themselves \cite{Harley2022}.
Doersch et al. \cite{Doersch2022} address this limitation by introducing the \texttt{TAP-Vid} benchmark consisting of labelled real and synthetic data for the TAP task, enabling the end-to-end point tracking model \texttt{TAP-Net}. 
The \texttt{TAPIR} method \cite{Doersch2023}  combines the advantages of both \texttt{TAP-Net} and \texttt{PIPs}.
Another enhancement is the \texttt{PIPs++} method by Zheng et al.~\cite{Zheng2023}. It extends \texttt{PIPs} for improved long-term tracking.
All the methods described above focus on tracking points independently, neglecting any potential correlations between them. 
Conversely, \texttt{OmniMotion} \cite{Wang2023} aims to achieve globally consistent motion by optimizing volumetric representations for individual videos. 
Similarly, \texttt{CoTracker} \cite{Karaev2023} tracks points jointly without relying on specific per-video optimization.
In this work, we employ a point tracking method to identify the location of points belonging to an occluded instance which is precisely what all aforementioned methods excel in. 
While any point tracking method can be used for our purpose, we choose to apply the \texttt{CoTracker} method \cite{Karaev2023} due to its performance and ease of use.

\noindent \textbf{Amodal Segmentation} describes the task of segmenting the full shape of an object even in the presence of occlusions. 
First methods investigate predicting an amodal mask given the visible mask and the input image \cite{Zhu2017a,Li2016a}. 
This has been extended to instance segmentation methods predicting the amodal instead of the visible mask directly from the input image \cite{Ling2020, Nguyen2021, Reddy2022, Sun2022,Follmann2019,Qi2019,Hu2019a}.
Amodal semantic segmentation methods apply grouping and multi-task training to look behind occlusions \cite{Purkait2019,Breitenstein2022,Breitenstein2022a}.
Recently, amodal video instance segmentation (VIS) has been investigated more extensively, considering both end-to-end supervised training of VIS \cite{Breitenstein2023}, and self-supervised approaches based on the visible masks \cite{Yao2022}. 
While the end-to-end approach requires amodal video labels which are expensive and difficult to obtain, the self-supervised approach requires visible masks to be predicted. 
In this sense, our approach is closest to the self-supervised approach, termed \texttt{SAVOS} \cite{Yao2022}, as we strive towards amodal VIS without relying on video-based labels.
\texttt{SAVOS} takes as input the images, the visible instance masks and the optical flow between two frames.
In contrast, we build our method on top of an image-based amodal segmentation method, where datasets for training are available. We rely on point tracking methods to track points of instances through the sequence which diminishes the need for optical flow. 
Moreover, while \texttt{SAVOS} makes its prediction in an offline fashion, i.e., based on the entire video, our proposed method can operate in an online fashion.

One challenge in amodal segmentation is the availability of real datasets, as amodal labeling of real data is associated with high costs, especially for videos. 
However, on image level, many datasets are available: The KINS dataset \cite{Qi2019} based on the KITTI dataset \cite{Geiger2013}, the COCOA dataset \cite{Zhu2017a} based on COCO \cite{Lin2014microsoft}, D2S \cite{Follmann2019}, KITTI-360-APS~\cite{Mohan2022} based on KITTI-360 \cite{Liao2021}, and Amodal Cityscapes \cite{Breitenstein2022} based on Cityscapes \cite{Cordts2016}. 
For videos, two large-scale synthetic video datasets with amodal video-level labels exist: SAIL-VOS \cite{Hu2019a} is a synthetic dataset collected in the GTA V game. AmodalSynthDrive \cite{Sekkat2023} is another synthetic dataset of automated driving sequences collected in the CARLA engine \cite{Dosovitskiy2017}. 
Additionally, on real data, Yao et al.~\cite{Yao2022} match amodal annotations of the KINS dataset with videos of KITTI to obtain amodal annotations on single frames of the videos, termed KINS-car. 
Note that in contrast to the synthetic datasets, for KINS-car no full video annotations are available. 
As we are interested in amodal video instance segmentation for automated driving, we investigate the performance of our method on AmodalSynthDrive and KINS-car. 
This way, we obtain evaluation results on videos. For training, only image annotations are needed.

    \vspace{-4pt}
\section{Proposed \texttt{S-AModal} Method}
\label{sec:method}
    \vspace{-2pt}

\begin{figure}[t]
    \centering
    \includegraphics[width=\textwidth]{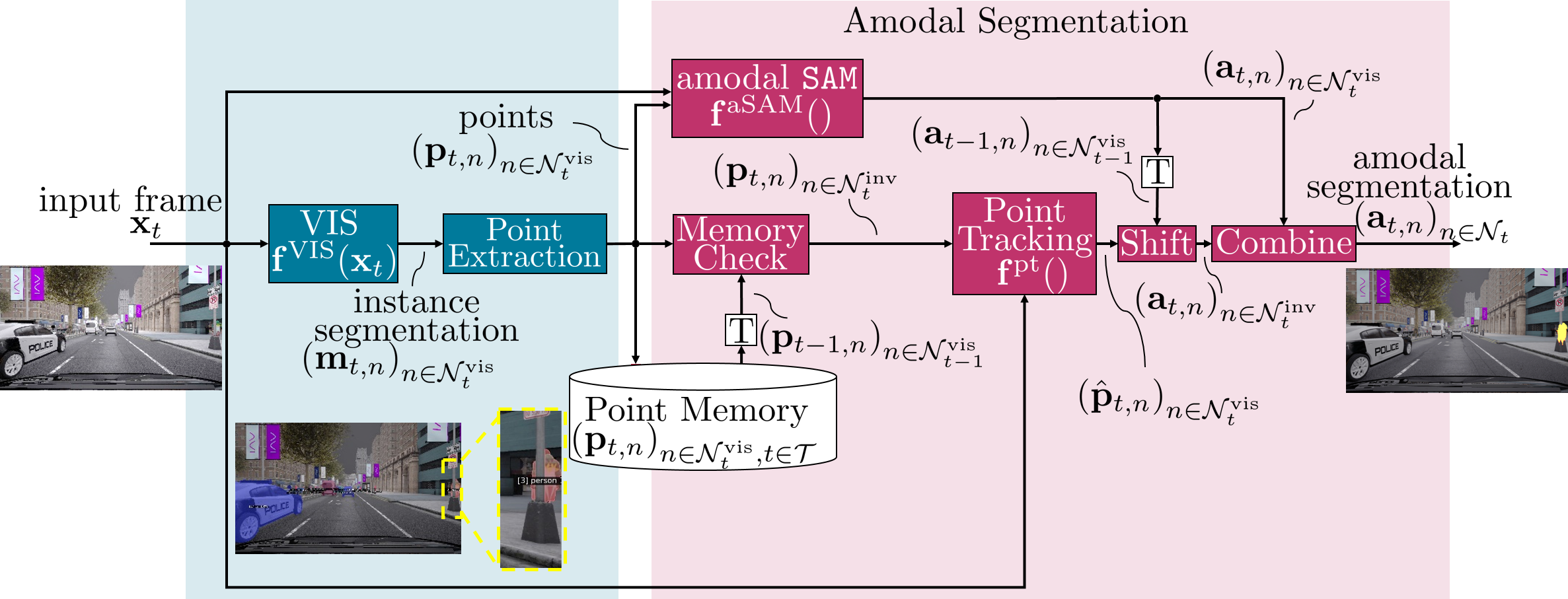}
    \caption{Overview of our \texttt{S-AModal} method: Given an input frame $\mathbf{x}_t$, it predicts amodal instance segmentation masks $\left(\mathbf{a}_{t,n}\right)_{n \in \mathcal{N}_{t}}$. First, a VIS method provides visible instance masks $\left(\mathbf{m}_{t,n}\right)_{n \in \mathcal{N}_{t}^\text{vis}}$, from which we extract points $\left(\mathbf{p}_{t,n}\right)_{n \in \mathcal{N}_{t}^\text{vis}}$.These points prompt our amodal SAM method to produce amodal instance masks $\left(\mathbf{a}_{t,n}\right)_{n \in \mathcal{N}_{t}^\text{vis}}$. Points are stored to track for occlusions, helping to update previous masks $\left(\mathbf{a}_{t-1,n}\right)_{n \in \mathcal{N}_{t-1}^\text{vis}}$ to $\left(\mathbf{a}_{t,n}\right)_{n \in \mathcal{N}_{t}^\text{inv}}$. Final amodal masks per frame $\left(\mathbf{a}_{t,n}\right)_{n \in \mathcal{N}_{t}}$ combine $\left(\mathbf{a}_{t,n}\right)_{n \in \mathcal{N}_{t}^\text{inv}}$ and $\left(\mathbf{a}_{t,n}\right)_{n \in \mathcal{N}_{t}^\text{vis}}$. We denote delay units by ``T''.}
    \label{fig:method}
        \vspace{-12pt}
\end{figure}

Our proposed \texttt{S-AModal} method performs amodal VIS based on point-prompting a fine-tuned \texttt{SAM} network which we term amodal \texttt{SAM} $\mathbf{f}^\text{aSAM}$. 
We consider a video as sequence of frames $\mathbf{x}_t \in \lbrack 0,1 \rbrack^{H \times W \times C}$ for $t \in \mathcal{T}=\lbrace 1, \ldots, T \rbrace$ with $T$ being the video length. 
The overall method is depicted in Figure \ref{fig:method}. 
We apply a standard online VIS method $\mathbf{f}^\text{VIS}$ to the input frame $\mathbf{x}_t$ to obtain the corresponding visible segmentation masks $\mathbf{f}^\text{VIS}(\mathbf{x}_t)=\left(\mathbf{m}_{t,n}\right)_{n \in \mathcal{N}_{t}^\text{vis}} \in \lbrace 0,1 \rbrace^{H\times W \times N_t}$ for all instances $n \in \mathcal{N}_{t}^\text{vis}= \lbrace1, \ldots, N_{t}^\text{vis} \rbrace$ with $N_{t}^\text{vis}$ being the number of instances visible in frame $t$. 
The total number of visible and occluded instances in frame $t$ is denoted as $N_t = N_t^\text{vis} + N_t^\text{inv}$, and the total number of instances per video sequence is accordingly denoted as $N$. 
We define a point selected from the instance segmentation $\mathbf{m}_{t,n}$ as $ p_{t,n,k} \in \lbrace 1, ..., H \cdot W \rbrace$, i.e., a pixel index where the visible instance mask is predicted. From each instance segmentation $\mathbf{m}_{t,n}$, we extract a point K-tuple (henceforth referred to as points) $\mathbf{p}_{t,n} = \left( p_{t,n,k} \right) \in \lbrace 1,..., H \cdot W \rbrace^{K}$ with $K$ being the number of points used to prompt the amodal \texttt{SAM} method $\mathbf{f}^\text{aSAM}$. 
The amodal \texttt{SAM} method then provides the amodal instance segmentation $\mathbf{f}^\text{aSAM}(\mathbf{x}_{t},\mathbf{p}_{t,n}) = \mathbf{a}_{t,n} \in \lbrace 0,1 \rbrace^{H \times W}$, $n \in \mathcal{N}_t^\text{vis}$. 
Note that both, point extraction and prompting of amodal \texttt{SAM}, are done for all visible instances with index $n \in \mathcal{N}_{t}^\text{vis}$.
Additionally, we store the extracted points $\mathbf{p}_{t,n}$ in a point memory, which we access for each frame to check for instances $n \in \mathcal{N}_t$ that are not visible (``memory check'' in Figure \ref{fig:method}). 
We term this subset of fully occluded (i.e., invisible) instances at frame $t$ as $\mathcal{N}_{t}^\text{inv}$. 
The point memory contains all previously observed point K-tuples per previously observed instance, i.e., $\left( \mathbf{p}_{\tau,n} \right)_{\tau \in \mathcal{T}_t,n \in \mathcal{N}_\tau}$ where $\mathcal{T}_t = \lbrace 1, \ldots, t \rbrace$, and $t$ being the current frame index. 
At frame $t$, a memory retrieval checks whether any points in the K-tuple $\mathbf{p}_{t-1,n}$ are contained in the memory, for which the corresponding instance $n$ has not been predicted by the VIS method $\mathbf{f}^\text{VIS}$ for frame $t$, i.e., $n \in \mathcal{N}_{t}^\text{inv}$.
A point tracking method $\mathbf{f}^\text{pt}$ then tracks the as occluded identified instance points $\left(\mathbf{p}_{t-1,n}\right)_{n \in \mathcal{N}_{t}^\text{inv}}$, towards the current frame $\mathbf{x}_{t}$, to obtain predicted points $\mathbf{f}^\text{pt}(\mathbf{x}_t, (\mathbf{p}_{t-1,n})_{n \in \mathcal{N}_{t}^\text{inv}} ) = \left(\hat{\mathbf{p}}_{t,n}\right)_{n \in \mathcal{N}_{t}^\text{inv}}$.

\begin{figure}[t]
    \centering
    \includegraphics[width=\textwidth]{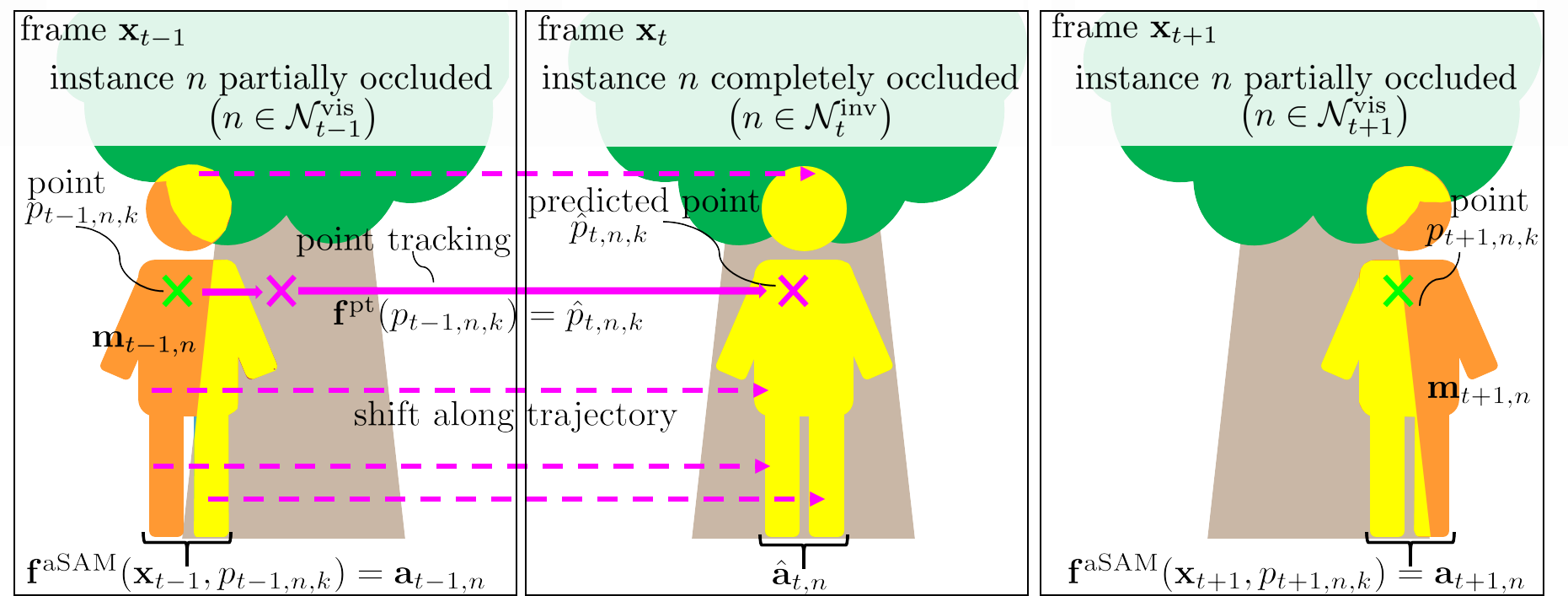}
    \caption{Schematic view of a video sequence $\mathbf{x}_{t-1}, \mathbf{x}_{t}, \mathbf{x}_{t+1}$ and an instance $n$ which is partially visible at $t-1$ at $t+1$, but fully occluded at $t$. For frames $\mathbf{x}_{t-1}$ and $\mathbf{x}_{t+1}$, the points $p_{t-1,n,k}$ and $p_{t+1,n,k}$ of the predicted visible instance masks $\mathbf{m}_{t-1,n}$ and $ \mathbf{m}_{t+1,n}$, are used to prompt the amodal \texttt{SAM} model to obtain $\mathbf{a}_{t-1,n}$ and $ \mathbf{a}_{t+1,n}$, respectively. For frame $\mathbf{x}_t$, we apply point tracking to obtain the predicted point $\hat{p}_{t,n,k}$ and shift the amodal mask along this trajectory to $\hat{\mathbf{a}}_{t,n}$.}
    \label{fig:schematic-view-point-tracking}
        \vspace{-12pt}
\end{figure}

Let's look into details: Given a point $p_{t-1,n,k} \in \lbrace 1,..., H \cdot W \rbrace$ corresponding to the 2D pixel position $(h_{t-1,n},w_{t-1,n})$, with $h_{t-1,n} \in \lbrace 1,\ldots,H \rbrace, w_{t-1,n} \in \lbrace1,\ldots,W\rbrace$, the goal of the point tracking method is to determine the pixel position of that point at the next time step $t$.
Moreover, a binary occlusion score $o_{t,n,k} \in \lbrace 0,1 \rbrace$ predicts whether the point is actually visible or occluded by another object.
Similar to the optical flow, these pixel positions can be represented as 2D $(\Delta h_{t-1,n}, \Delta w_{t-1,n})$ displacement vectors at time step $t-1$, where $\Delta h_{t-1,n} \in \mathbb{R}$ describes the vertical offset with respect to the source position and $\Delta w_{t-1,n} \in \mathbb{R}$ represents the horizontal offset.
In summary, for each point $p_{t-1,n,k}$ that should be tracked, a point tracking method $\mathbf{f}^\text{pt}$ predicts the point $\hat{p}_{t,n,k}$ according to $\mathbf{f}^\text{pt}(\mathbf{x}_t, (\mathbf{p}_{t-1,n}))=(\hat{\mathbf{p}}_{t,n},)$ and the occlusion score $o_{t,n,k}$. 
From the predicted point $\hat{p}_{t,n,k}$, we calculate vertical offset $\Delta h_{t-1,n}$ and the horizontal offset $\Delta w_{t-1,n}$, which we then use to shift the amodal mask $\mathbf{a}_{t-1,n}$ to time step $t$, i.e., $\mathbf{a}_{t,n}$.
This way we obtain a trajectory from the point $p_{t-1,n,k}$, $n \in \mathcal{N}_{t}^\text{inv}$, to its position in the next frame, i.e., $\hat{p}_{t,n,k}$. 

Figure \ref{fig:schematic-view-point-tracking} shows this for a schematic example of an occluded person. 
In image frame $\mathbf{x}_{t-1}$, the person is partially visible and amodal \texttt{SAM} $\mathbf{f}^\text{aSAM}$ is prompted using a point $p_{t-1,n,k}$ (green) from the visible mask $\mathbf{m}_{t-1,n}$ (orange) to obtain the full amodal mask for instance $n$, i.e., $\mathbf{f}^\text{aSAM}(\mathbf{x}_{t-1},p_{t-1,n,k}) = \mathbf{a}_{t-1,n}$.
Using a VIS method for tracking and segmentation allows identifying a fully occluded instance as a missing segmentation mask in the output.
As this instance $n$ is no longer visible in frame $\mathbf{x}_t$, we use a point tracking method $\mathbf{f}^\text{pt}$ to predict the location of $p_{t-1,n,k}$ in $\mathbf{x}_t$, i.e., obtaining $\mathbf{f}^\text{pt}(\mathbf{x}_t, p_{t-1,n,k})=\hat{p}_{t,n,k}$ (illustrated in pink in Figure \ref{fig:schematic-view-point-tracking}). 
This gives a trajectory as visualized by pink arrows in Figure \ref{fig:schematic-view-point-tracking}. 
The amodal mask is moved along this trajectory to obtain the amodal mask $\hat{\mathbf{a}}_{t,n}$.
If the number of points in the K-tuple is $K>1$, we obtain a trajectory per point from the point tracking method. 

We use the predicted points $\left(\hat{\mathbf{p}}_{t,n}\right)_{n \in \mathcal{N}_{t}^\text{inv}}$ to shift the corresponding previously observed amodal instance masks $\left(\hat{\mathbf{a}}_{t-1,n}\right)_{n \in \mathcal{N}_{t}^\text{inv}}$ to the current frame $\mathbf{x}_t$ along the trajectory, and obtain the amodal masks for instances that were not detected by the VIS method in frame $\mathbf{x}_t$, i.e., $\left(\hat{\mathbf{a}}_{t,n}\right)_{n \in \mathcal{N}_{t}^\text{inv}}$.
The simple shifting operation translates the previous amodal mask $\mathbf{a}_{t-1,n}$ to the current frame by adding the displacement from the previous point $p_{t-1,n,k}$ to the predicted point $\hat{p}_{t,n,k}$ to all coordinates of the amodal mask $\mathbf{a}_{t-1,n}$. Note that if $K>1$, we calculate the displacement as the average of the predicted points.  
Combining the amodal masks from the point tracking branch $\left(\hat{\mathbf{a}}_{t,n}\right)_{n \in \mathcal{N}_{t}^\text{inv}}$ and the amodal \texttt{SAM} branch $\left(\hat{\mathbf{a}}_{t,n}\right)_{n \in \mathcal{N}_{t}^\text{vis}}$ gives the full set of amodal instance masks for frame $\mathbf{x}_t$, $\left(\hat{\mathbf{a}}_{t,n}\right)_{n \in \mathcal{N}_{t}}$.

Our approach is independent of the choice of the VIS method $\mathbf{f}^\text{VIS}$ and of the point tracking method $\mathbf{f}^\text{pt}$, so any method in these fields can be chosen. Our amodal \texttt{SAM} is fine-tuned on the amodal ground truth on images. For fine-tuning \texttt{SAM} on amodal data, we follow the approach of Chen et al.~\cite{Chen2023}, keeping the image encoder fixed, while we add additional adapter layers to the image encoder and fine-tune the decoder. In addition, we keep the fixed prompt encoder to insert point prompts into the network, see more details in the Supplementary Sec. A. 

    \vspace{-4pt}
\section{Experimental Setup}
\label{sec:experimental-setup}
    \vspace{-2pt}

We consider two datasets: AmodalSynthDrive $\mathcal{D}_\text{ASD}$ \cite{Sekkat2023} and KINS-car $\mathcal{D}_\text{Kcar}$ \cite{Yao2022}. As $\mathcal{D}_\text{ASD}$ only provides training ($\mathcal{D}^\text{train}_\text{ASD}$) and validation ($\mathcal{D}^\text{val}_\text{ASD}$) data with ground-truth annotations, we report results on the validation set. For $\mathcal{D}_\text{Kcar}$, in addition to training ($\mathcal{D}^\text{train}_\text{Kcar}$) and validation ($\mathcal{D}^\text{val}_\text{Kcar}$) data, labeled test data $\mathcal{D}^\text{test}_\text{Kcar}$ exists on which our results are reported.
Both are video-based datasets, while $\mathcal{D}_\text{Kcar}$ only provides an image-based ground truth and no tracking information. 

On \textit{image level}, we report mean intersection over union (mIoU), average precision (AP), and AP$_{50}$ according to the COCO evaluation \cite{Lin2014microsoft}. 
Additionally, we report derivatives of these metrics for small, medium, and large objects as well as partial and heavy occlusions following standards in literature \cite{Lin2014microsoft,Hu2019a,Breitenstein2023}.

On \textit{video level}, we report video average precision (vAP) and its derivatives metrics following the SAIL-VOS dataset benchmark \cite{Hu2019a}, and Breitenstein et al.~\cite{Breitenstein2023}. Definitions are recapitulated in the Supplementary Sec.\ B. 
Note that for our proposed method, the instance class prediction is simply derived from the underlying VIS method. Hence, all our metrics are reported in a class-agnostic setting. The quality of class prediction is not influenced by the investigated methods and gives no indication about their performance. In this setting, class predictions do not influence whether a prediction is defined as true positive, false positive or false negative, instead, this decision is only based on the IoU between predicted and ground-truth mask.

We compare our results with \texttt{SAVOS} \cite{Yao2022}, which is closest to our method as it also does not require amodal video-based labels during training. 
\texttt{SAVOS} performs self-supervised amodal VIS on top of a visible instance segmentation.
It is trained without amodal labels using optical flow and the visible instance masks.  
We train \texttt{SAVOS} \cite{Yao2022} following the original setting on both datasets.

On $\mathcal{D}_\text{Kcar}$, we use \texttt{PointTrack} (\texttt{PT}) \cite{Xu2020} as VIS method with the same checkpoint as \texttt{SAVOS} \cite{Yao2022}. 
For $\mathcal{D}_\text{ASD}$, we choose \texttt{GenVIS} \cite{Heo2023}, one of the current top VIS methods. We train \texttt{GenVIS} on $\mathcal{D}_\text{ASD}^\text{train}$ using the full image resolution. It achieves an AP of $30.38\%$, an AP$_{50}$ of $43.83\%$, a vAP of $16.36\%$ and a vAP$_{50}$ of $23.13\%$ on $\mathcal{D}_\text{ASD}^\text{val}$. Detailed results are reported in Supplementary Sec.\ C. We also replace the VIS method in Figure \ref{fig:method} by the ground truth (GT) to cancel out the VIS performance, giving the methods access to the ground-truth visible masks.

To obtain our amodal \texttt{SAM} $\mathbf{f}^\text{aSAM}$, \texttt{SAM} with the added adapters is fine-tuned on the training data of both datasets using a batchsize of 1 on one \texttt{NVidia A100} GPU for 20 epochs.
Due to memory constraints, we use the vision transformer \texttt{ViT-B} backbone \cite{Dosovitskiy2021} with pre-trained weights as provided by the original \texttt{SAM} codebase \cite{Kirillov2023}.
We use the AdamW optimizer and a start learning rate of 0.00001.
The learning rate is multiplied by $0.1$ every $10$ epochs. 
As loss function, we use a combination of focal loss and dice loss as is common for fine-tuning \texttt{SAM} \cite{Chen2023,Shaharabany2023,Wu2023}. We report results as mean across three inference runs. If not stated otherwise, we prompt amodal \texttt{SAM} $\mathbf{f}^\text{aSAM}$ with $K=1$ point.
As point tracking method, we use \texttt{CoTracker} \cite{Karaev2023} with the given checkpoints. No further fine-tuning is necessary.

\vspace{-4pt}
\section{Experiments and Discussion}
\label{sec:experiments}
    \vspace{-2pt}


\begin{table*}[t!]
\footnotesize
\def\arraystretch{0.8}%
\setlength{\tabcolsep}{2.5pt}
    \centering
    \begin{tabular}{c|c|c|c|c|c|c|c|c|c}
        Data\!  & Method\! & VIS  &  AP &  AP$_{50}$ & AP$^\text{P}_{50}$ & AP$^\text{H}_{50}$  &  AP$^\text{L}_{50}$ &AP$^\text{M}_{50}$ & AP$^\text{S}_{50}$ \\
        \hline \hline

                        \stz \multirow{4}{*}{$\mathcal{D}^\text{val}_\text{ASD}$}  & \texttt{SAVOS} \cite{Yao2022} & \texttt{GV} & \textcolor{black}{\phantom{0}$7.41$} & \textcolor{black}{$13.19$} & \textcolor{black}{$16.13$}  & \textcolor{black}{\phantom{0}$2.34$} & \textcolor{black}{$12.52$}& \textcolor{black}{$16.24$} & \textcolor{black}{\phantom{0}$6.91$} \\ 
        \stz & \texttt{S-AModal} (ours)& \texttt{GV}  & \textcolor{black}{$\mathbf{21.59}$} & \textcolor{black}{$\mathbf{35.00}$} & \textcolor{black}{$\mathbf{38.50}$}  & \textcolor{black}{$\mathbf{10.43}$}  & \textcolor{black}{$\mathbf{66.71}$} & \textcolor{black}{$\mathbf{50.23}$} & \textcolor{black}{$\mathbf{14.75}$}   \\ 
        \cline{2-10}         
        \stz & \texttt{SAVOS} \cite{Yao2022}& GT  & $\mathbf{50.89}$ & $\mathbf{76.26}$ & $\mathbf{81.89}$  & ${41.92}$  & ${91.49}$ & $\mathbf{84.21}$ & ${41.62}$   \\ 
                  \stz & \texttt{S-AModal} (ours) & GT & $46.91$ & $73.86$ & $80.85$  & $\mathbf{43.79}$  & $\mathbf{97.15}$ & ${82.41}$ & $\mathbf{60.13}$   \\ 
        \hline
        \stz \multirow{2}{*}{ $\mathcal{D}^\text{test}_\text{Kcar}$} & \texttt{SAVOS} \cite{Yao2022} & \texttt{PT} & $40.50$ & $61.80$ & $77.89$ & $30.72$ & $96.38$ & $\mathbf{89.83}$ & $39.38$ \\
                \stz  & \texttt{S-AModal} (ours)& \texttt{PT}& $\mathbf{41.08}$  & $\mathbf{74.28}$  & $\mathbf{78.23}$ & $\mathbf{35.08}$ & $\mathbf{97.40}$ & $85.25$ & $\mathbf{57.13}$ \\ 
     
    \end{tabular}
    \vspace{5pt}
    \caption{Amodal \textbf{image-level} instance segmentation metrics on $\mathcal{D}^\text{val}_\text{ASD}$ and on $\mathcal{D}^\text{test}_\text{Kcar}$ using \texttt{GenVIS} (\texttt{GV}) \cite{Heo2023}, the ground-truth visible masks (GT), and \texttt{PointTrack} (\texttt{PT}) \cite{Xu2020} as VIS methods, respectively. Best results in \textbf{bold}.}
    \label{tab:image-results-samadpt}
    \vspace{-20pt}
\end{table*}

\noindent \textbf{Quantitative Results:} First, we regard the amodal image-level results on both datasets. 
Table \ref{tab:image-results-samadpt} reports the results on the validation data of {AmodalSynthDrive} ($\mathcal{D}^\text{val}_\text{ASD}$) and on the test data of {KINS-car} ($\mathcal{D}^\text{test}_\text{Kcar}$). 
We compare our results against the \texttt{SAVOS} method \cite{Yao2022}. 
We observe that our \texttt{S-AModal} method outperforms \texttt{SAVOS} on both datasets in almost all metrics. 
Especially, on $\mathcal{D}^\text{test}_\text{Kcar}$ with the \texttt{PointTrack} (\texttt{PT}) method \cite{Xu2020}, the average precision (AP) of \texttt{S-AModal} on heavily occluded objects (AP$^\text{H}_{50}$), \texttt{S-AModal} ($35.08\%$) excels the \texttt{SAVOS} result ($30.72\%$) by $4.36\%$ absolute.
\textcolor{black}{We observe similar performance improvements on $\mathcal{D}^\text{val}_\text{ASD}$ using the VIS method \texttt{GenVIS} \cite{Heo2023}: \texttt{S-AModal} leads to better performance in comparison to \texttt{SAVOS}, even increasing AP by $14.18\%$ absolute to $21.59\%$ and AP$^\text{H}_{50}$ by $21.81\%$ absolute to $35.00\%$.}
We perform an additional experiment on $\mathcal{D}^\text{val}_\text{ASD}$, using the visible ground truth as VIS method. Table \ref{tab:image-results-samadpt} shows the results: 
Especially for heavy occlusions \texttt{S-AModal} increases AP$^\text{H}_{50}$ by $1.87\%$ absolute to $43.79\%$. 
However, \texttt{SAVOS} achieves slightly better results on 4 out of the 7 metrics, e.g., AP and AP$_{50}$. This can be attributed to the differences in both methods which gives \texttt{SAVOS} an advantage in this setting: \texttt{SAVOS} takes as input the image, the visible mask and the optical flow to predict the full amodal mask. This means that in the ground truth setting it can simply learn to just add amodal areas beside the visible mask. In contrast, \texttt{S-AModal} only takes points of the visible mask as input to predict the amodal mask, and hence, cannot leverage the full mask-specific information, such as shape. However, the improvement in AP$^\text{H}_{50}$ shows that once amodal and visible mask become more different due to occlusion, our proposed approach is better suited for this task. The higher robustness of our approach for practical cases, where no ground truth (GT) is available, can be seen in the first and the third row segment of Table \ref{tab:image-results-samadpt}.

\begin{table*}[t!]
\footnotesize
\def\arraystretch{0.8}%
\setlength{\tabcolsep}{2.5pt}
    \centering
    \begin{tabular}{c|c|c|c|c|c|c|c|c|c}
        Data & Method  & VIS & vAP &  vAP$_{50}$ & vAP$^\text{P}_{50}$ & vAP$^\text{H}_{50}$  &  vAP$^\text{L}_{50}$ & vAP$^\text{M}_{50}$ & vAP$^\text{S}_{50}$ \\ 
        \hline \hline
                        \stz \multirow{4}{*}{$\mathcal{D}^\text{val}_\text{ASD}$} & \texttt{SAVOS} \cite{Yao2022} & \texttt{GV} & \textcolor{black}{\phantom{0}$1.27$} & \textcolor{black}{\phantom{0}$3.95$} & \textcolor{black}{$12.46$}  & \textcolor{black}{\phantom{0}$0.16$} & \textcolor{black}{\phantom{0}$7.56$}& \textcolor{black}{\phantom{0}$0.16$} & \textcolor{black}{\phantom{0}$0.43$} \\

        \stz  & \texttt{S-AModal} (ours) & \texttt{GV}  & \textcolor{black}{\phantom{0}$\mathbf{2.96}$} & \textcolor{black}{\phantom{0}$\mathbf{6.03}$} & \textcolor{black}{$\mathbf{13.44}$}  & \textcolor{black}{\phantom{0}$\mathbf{0.63}$}  & \textcolor{black}{$\mathbf{27.91}$} & \textcolor{black}{\phantom{0}$\mathbf{4.12}$} & \textcolor{black}{\phantom{0}$\mathbf{1.64}$}   \\
        \cline{2-10}
                \stz  & \texttt{SAVOS} \cite{Yao2022} & GT & {20.53} & {38.39} & {45.54} & {31.39} & {79.60} & {73.46} & {14.77}  \\
        \stz  & \texttt{S-AModal} (ours) & GT &$\textcolor{black}{\mathbf{30.88}}$ & $\textcolor{black}{\mathbf{56.60}}$   & $\textcolor{black}{\mathbf{67.03}}$  & $\textcolor{black}{\mathbf{45.57}}$  & $\textcolor{black}{\mathbf{98.91}}$  & $\textcolor{black}{\mathbf{80.46}}$   & $\textcolor{black}{\mathbf{29.10}}$  \\ 
    \end{tabular}
    \vspace{5pt}
    \caption{Amodal \textbf{video-level} instance segmentation metrics on $\mathcal{D}^\text{val}_\text{ASD}$ using \texttt{GenVIS} (\texttt{GV}) \cite{Heo2023} and ground-truth visible masks (GT) as VIS methods. Best results in \textbf{bold}.}
    \label{tab:video-results-asd-amodal}
    \vspace{-20pt}
\end{table*}

\begin{figure}[t]
    \centering
        \begin{minipage}[t]{0.02\textwidth}
        \rotatebox{90}{\hspace{10pt} Video 1}
        \rotatebox{90}{\hspace*{-1.4cm} Video 2}
        \rotatebox{90}{\hspace*{-3.1cm} Video 3}
    \end{minipage}
\begin{minipage}[t]{0.95\textwidth}
    \centering
    \includegraphics[width=0.32\textwidth]{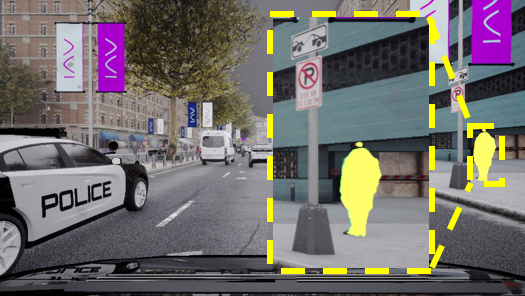}
        \includegraphics[width=0.32\textwidth]{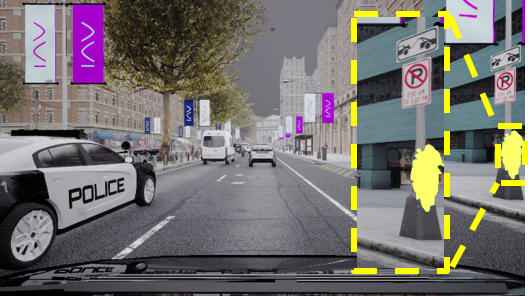}
            \includegraphics[width=0.32\textwidth]{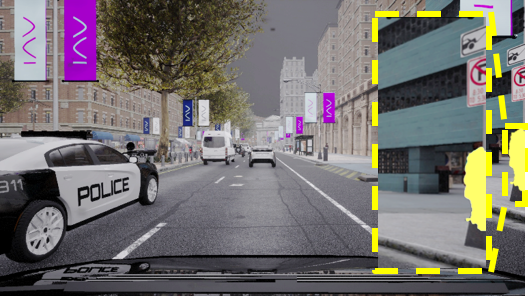}
                \includegraphics[width=0.32\textwidth]{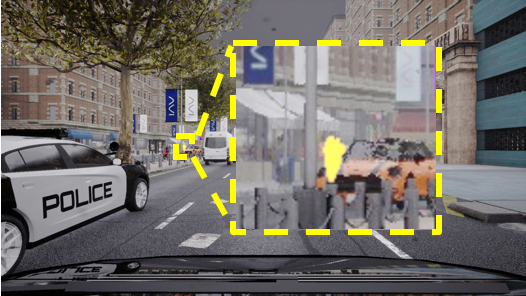}
    \includegraphics[width=0.32\textwidth]{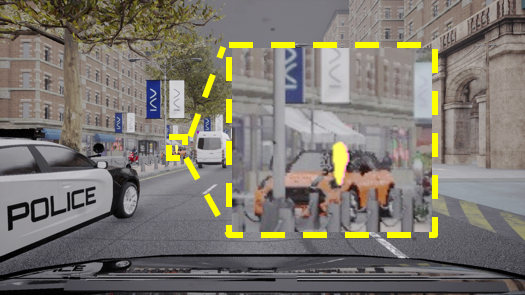}
        \includegraphics[width=0.32\textwidth]{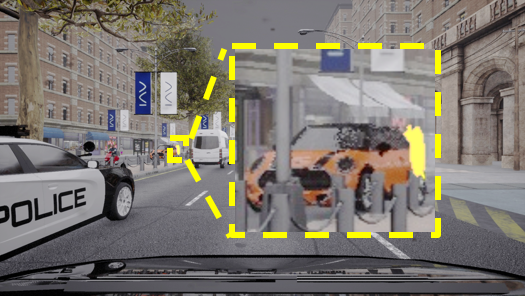}
            \includegraphics[width=0.32\textwidth]{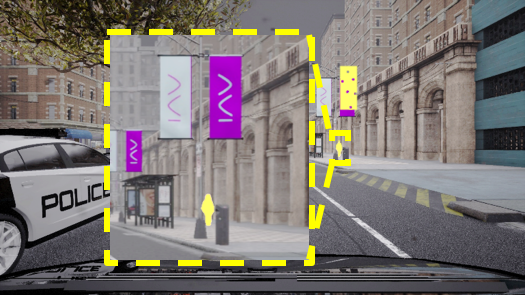}
                \includegraphics[width=0.32\textwidth]{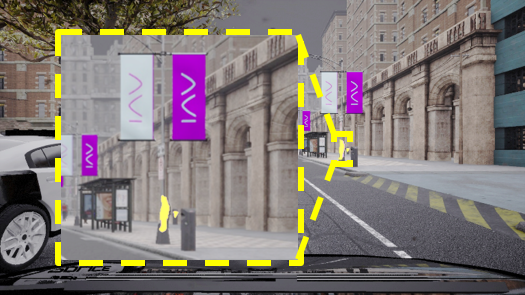}    
                        \includegraphics[width=0.32\textwidth]{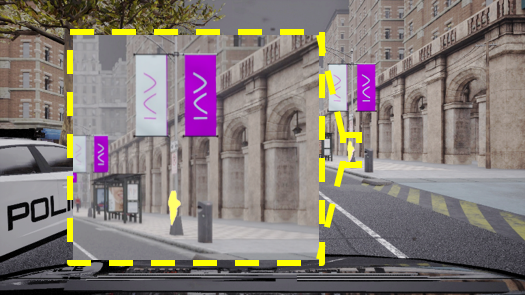}
                \end{minipage}
\begin{minipage}[t]{0.95\textwidth}
    \centering
$t-1$ \; \; \; \;\; \; \;\; \; \; \; \;\; \; \; \; \;\; \  $t$ \ \; \; \;\; \; \;\;\; \; \; \; \; \; \; \; \; \; \; $t+1$
\end{minipage}
\vspace{-7pt}
                \caption{\textbf{Qualitative results} of the proposed \texttt{S-AModal} method for three sequences $\mathbf{x}_{t-1}^{t+1}$ with overlayed colorized amodal predictions $\mathbf{a}_{t-1}^{t+1}$ on AmodalSynthDrive $\mathcal{D}^\text{val}_\text{ASD}$.}
    \label{fig:qualitative results-asd}
        \vspace{-12pt}
\end{figure}

\begin{figure}[ht!]
    \centering
\begin{minipage}[t]{0.99\textwidth}
    \centering
    \includegraphics[width=0.32\textwidth]{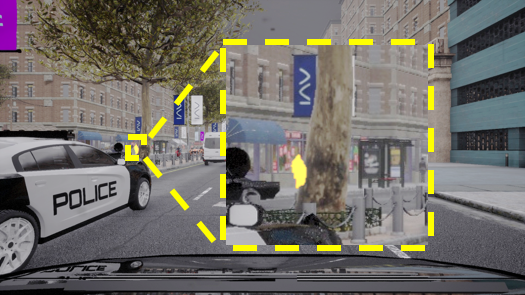}
        \includegraphics[width=0.32\textwidth]{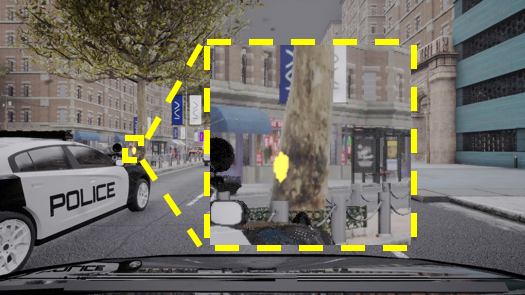}
                         \end{minipage}
                      
        \begin{minipage}[t]{0.95\textwidth}
    \centering
    \vspace*{-8pt}       
$t-3$ \; \; \; \;\; \; \;\; \; \; \; \;\; \; \; \; \;\; \  $t-2$ 
\end{minipage}      

\begin{minipage}[t]{0.95\textwidth}   \includegraphics[width=0.32\textwidth]{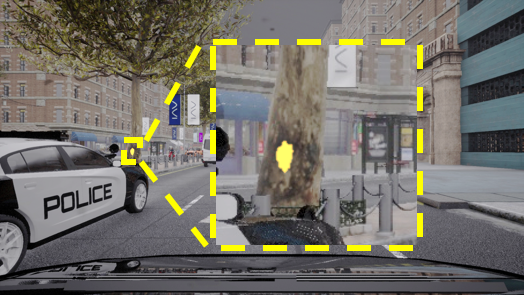}
                \includegraphics[width=0.32\textwidth]{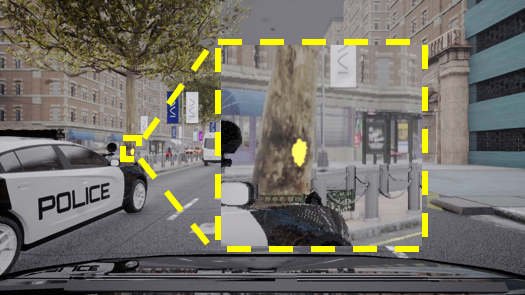}
    \includegraphics[width=0.32\textwidth]{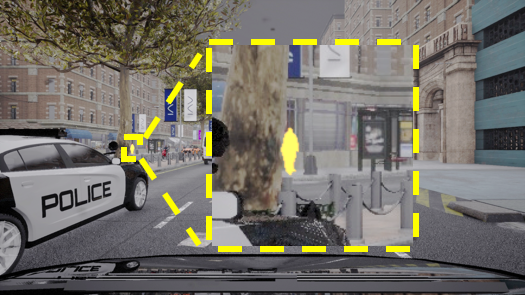}
                \end{minipage}
\begin{minipage}[t]{0.95\textwidth}
    \centering
    \vspace*{-8pt}    
$t-1$ \; \; \; \;\; \; \;\; \; \; \; \;\; \; \; \; \;\; \  $t$ \ \; \; \;\; \; \;\;\; \; \; \; \; \; \; \; \; \; \; $t+1$
\end{minipage}
\vspace{-7pt}
                \caption{\textbf{Qualitative results} of the proposed \texttt{S-AModal} method for a sequence $\mathbf{x}_{t-3}^{t+1}$ with overlayed colorized amodal predictions $\mathbf{a}_{t-3}^{t+1}$ on AmodalSynthDrive $\mathcal{D}^\text{val}_\text{ASD}$, illustrating a fully occluded person for a time span of 3 frames ($t-2,t-1,t$).}
    \label{fig:qualitative results-asd-fullsequence}
        \vspace{-12pt}
\end{figure}

Table \ref{tab:video-results-asd-amodal} reports the video-level metrics of the proposed \texttt{S-AModal} method and \texttt{SAVOS} \cite{Yao2022} on $\mathcal{D}^\text{val}_\text{ASD}$. 
Note that we cannot report video-level results for $\mathcal{D}^\text{test}_\text{Kcar}$ since video-level ground truth is not available. 
Using \texttt{GenVIS}, we see that results on video level show the same tendency as on image level: \texttt{S-AModal} outperforms \texttt{SAVOS} in all metrics, e.g., leading to an \textcolor{black}{$1.69\%$} absolute performance improvement towards a vAP$_{50} $ of $ 2.96\%$. 
The amodal VIS results of Table \ref{tab:video-results-asd-amodal} are in line with the VIS results of \texttt{GenVIS} and can be attributed to relatively low tracking performance of the underlying VIS method. 
When using the ground truth as input (lower segment in Table \ref{tab:video-results-asd-amodal}), \texttt{S-AModal} interestingly outperforms \texttt{SAVOS} in all metrics, leading to an $14.18\%$ absolute improvement in vAP$^\text{H}_{50}$, showing that \texttt{S-AModal} is better suited to handle heavy occlusions. 


\noindent \textbf{Qualitative Results:} Figure \ref{fig:qualitative results-asd} shows qualitative results of \texttt{S-AModal} for three sequences $\mathbf{x}_{t-1}^{t+1}$ of the validation data of AmodalSynthDrive $\mathcal{D}^\text{val}_\text{ASD}$. The amodal predictions $\mathbf{a}_{t-1}^{t+1}$ are overlayed over the image for visualization purposes. The same color indicates the same identified instance. In all three videos, occluded pedestrians are tracked through occlusions with plausible amodal masks showing that our \texttt{S-AModal} method provides high-quality results in amodal VIS.

Figure \ref{fig:qualitative results-asd-fullsequence} shows an example of \texttt{S-AModal} with a longer occlusion of a person (yellow mask) behind a large tree spanning 3 frames ($t-2,t-1,t$). The amodal mask of the last appearance at $t-3$ is shifted along the predicted point trajectory to frames $t-2$, $t-1$, $t$. \texttt{S-AModal} is clearly able to follow the person throughout this full occlusion until its reappearance in frame $t+1$.

\begin{figure}[t]
    \centering
    \begin{minipage}[t]{0.02\textwidth}
        \rotatebox{90}{\hspace{10pt} \texttt{SAVOS} \cite{Yao2022}}
        \rotatebox{90}{\hspace*{-1.6cm} \texttt{S-SAModal}}
    \end{minipage}
\begin{minipage}[t]{0.96\textwidth}
    \centering
    \includegraphics[width=0.32\textwidth]{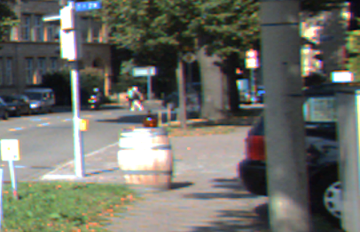}
        \includegraphics[width=0.32\textwidth]{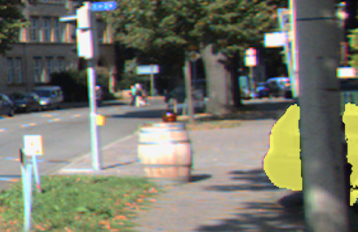}
            \includegraphics[width=0.32\textwidth]{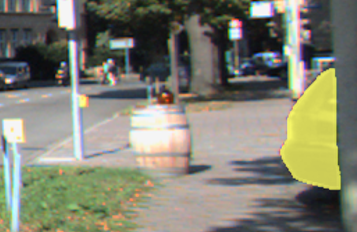}
                \includegraphics[width=0.32\textwidth]{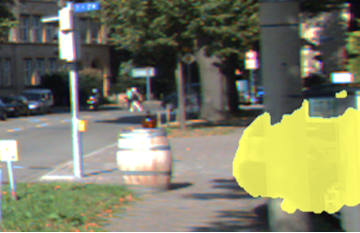}
    \includegraphics[width=0.32\textwidth]{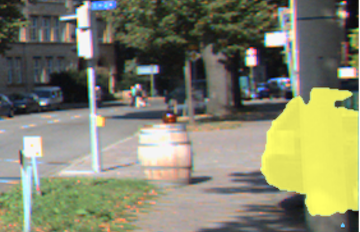}
        \includegraphics[width=0.32\textwidth]{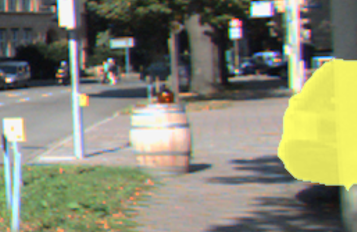}
                \end{minipage}
                \begin{minipage}[t]{0.95\textwidth}
    \centering
$t-1$ \; \; \; \;\; \; \;\; \; \; \; \;\; \; \; \; \;\; \  $t$ \ \; \; \;\; \; \;\;\; \; \; \; \; \; \; \; \; \; \; $t+1$
\end{minipage}
\vspace{-7pt}
                \caption{\textbf{Qualitative results} of the \texttt{SAVOS} \cite{Yao2022} (top) and the proposed \texttt{S-AModal} (bottom) methods for a sequence $\mathbf{x}_{t-1}^{t+1}$ with yellow amodal predictions $\mathbf{a}_{t-1}^{t+1}$ on $\mathcal{D}^\text{test}_\text{Kcar}$.}
    \label{fig:qualitative results-kcar}
      \vspace{-10pt}
\end{figure}

\textcolor{black}{In Figure \ref{fig:qualitative results-kcar} we compare our proposed \texttt{S-AModal} method (bottom) with the \texttt{SAVOS} method (top) on a sequence $\mathbf{x}_{t-1}^{t+1}$ of $\mathcal{D}^\text{test}_\text{Kcar}$. The quantitative performance gain of \texttt{S-AModal} is reflected in these qualitative results as well. The visualized amodal segmentation masks of the car (yellow) are seemingly more accurate compared to the \texttt{SAVOS} results in handling the occlusion by the tree. Note that the tracking quality of both methods is inherited from \texttt{PointTrack} \cite{Xu2020}. When the VIS method fails as in frame $t-1$, \texttt{SAVOS} is not able to make a prediction.}

\begin{table*}[t!]
\footnotesize
\def\arraystretch{0.8}%
\setlength{\tabcolsep}{1.5pt}
    \centering
    \begin{tabular}{c|r|r|c|c}
    & \multicolumn{4}{c}{Number of Point Prompts} \\
       & \multicolumn{1}{c|}{$K=1$} & \multicolumn{1}{c|}{$K=2$} & $K=3$& $K=4$\\
              \hline \hline  
        \stzdown AP\phantom{$_{50}$} & $\mathbf{46.91} \pm 0.03$ & $41.75 \pm 0.03$ & $35.47 \pm 0.01$ & $30.01 \pm 0.13$ \\
        
        AP$_{50}$ & $\mathbf{73.86} \pm 0.06$ & $68.57 \pm 0.21$ & $61.20 \pm 0.09$ & $53.54 \pm 0.27$ \\
        
        AP$^\text{P}_{50}$ & $\mathbf{80.85} \pm 0.08$ & $76.45 \pm 0.23$ & $69.11 \pm 0.14$ & $62.34 \pm 0.07$ \\
        
        AP$^\text{H}_{50}$  &  $\mathbf{43.79} \pm 0.38$ & $37.58 \pm 0.61$ & $28.36 \pm 0.08$ & $20.53 \pm 0.11$ \\
        
        AP$^\text{L}_{50}$ & ${97.15} \pm 0.53$ & $\mathbf{97.71} \pm 0.42$ & $96.84 \pm 0.15$ & $95.94 \pm 0.16$ \\
        
        AP$^\text{M}_{50}$ & $\mathbf{82.41} \pm 0.08$ & $80.70 \pm 0.62$ & $72.56 \pm 0.11$ & $63.23 \pm 0.51$ \\
        
        AP$^\text{S}_{50}$ & $\mathbf{60.13} \pm 0.42$ & $51.39 \pm 0.15$ & $42.88 \pm 0.11$ & $34.95 \pm 0.01$ \\
    \end{tabular}
    \vspace{5pt}
    \caption{Ablation: Amodal \textbf{image-level} instance segmentation metrics in AP and its metric derivatives on $\mathcal{D}^\text{val}_\text{ASD}$ using the ground-truth visible masks (GT) as VIS method and using different numbers of points $K$ for prompting the amodal \texttt{SAM} network of \texttt{S-AModal}. Best results in \textbf{bold}.}
    \label{tab:ablation-results-point-prompts}
        \vspace{-20pt}
\end{table*}

\noindent \textbf{Ablation:} Amodal \texttt{SAM} is prompted using points as input. For all above experiments the number of point prompts per instance was chosen as one. 
Table \ref{tab:ablation-results-point-prompts} shows the results of the proposed \texttt{S-AModal} on image-level using different numbers $K$ of points to prompt the amodal \texttt{SAM} network. Surprisingly, using more points does not lead to better performance. Using only $K=1$ point leads to the best performance overall, e.g., an AP of 46.91\%. 
Only for large objects, AP$^\text{L}_{50}$ is slightly higher ($97.71\%$) when using $K=2$ point prompts compared to $K=1$. 
However, when regarding the standard deviations, the slightly higher mean value is not significant. 
Since our point prompts are randomly selected from the visible mask, additional points may not provide additional information to amodal \texttt{SAM}. 

\begin{table*}[t!]
\footnotesize
\def\arraystretch{0.8}%
\setlength{\tabcolsep}{1.5pt}
    \centering
    \begin{tabular}{c|c|c|c|c}
    & \multicolumn{4}{c}{Point Selection Method } \\
       & Random & Saliency & Erosion (default) & Erosion (best) \\
              \hline \hline  
        \stzdown AP\phantom{$_{50}$} & ${46.91} \pm 0.03$ & $36.84 \pm 0.15$ & $48.03 \pm 0.18$ & $\mathbf{49.21} \pm 0.04$ \\
        
        AP$_{50}$ & ${73.86} \pm 0.06$ & $58.16 \pm 0.19$ & $75.25 \pm 0.10$ & $\mathbf{76.26} \pm 0.38$ \\
        
        AP$^\text{P}_{50}$ & ${80.85} \pm 0.08$ & $65.83 \pm 0.28$ & $82.30 \pm 0.10$ & $\mathbf{82.85} \pm 0.07$ \\
        
        AP$^\text{H}_{50}$  &  ${43.79} \pm 0.38$ & $22.87 \pm 0.31$ & $45.90 \pm 0.25$ & $\mathbf{47.84} \pm 0.24$ \\
        
        AP$^\text{L}_{50}$ & ${97.15} \pm 0.53$ & ${88.76} \pm 0.53$ & $97.65 \pm 0.45$ & $\mathbf{98.05} \pm 0.11$ \\
        
        AP$^\text{M}_{50}$ & ${82.41} \pm 0.08$ & $69.15 \pm 0.72$ & $83.58 \pm 0.50$ & $\mathbf{85.61} \pm 0.16$ \\
        
        AP$^\text{S}_{50}$ & ${60.13} \pm 0.42$ & $42.17 \pm 0.08$ & $62.17 \pm 0.09$ & $\mathbf{62.66} \pm 0.43$ \\
    \end{tabular}
    \vspace{5pt}
    \caption{Ablation: Amodal \textbf{image-level} instance segmentation metrics in AP and its metric derivatives on $\mathcal{D}^\text{val}_\text{ASD}$ using the ground-truth visible masks (GT) as VIS method and using different point selection methods to prompt the amodal \texttt{SAM} network of \texttt{S-AModal}. Best results in \textbf{bold}.}
    \label{tab:ablation-point-selection}
        \vspace{-25pt}
\end{table*}

\begin{figure}[t]
    \centering
        \begin{minipage}[t]{0.02\textwidth}
        \rotatebox{90}{\hspace{10pt} Video 1}
        \rotatebox{90}{\hspace*{-1.4cm} Video 2}
        \rotatebox{90}{\hspace*{-3.1cm} Video 3}
    \end{minipage}
\begin{minipage}[t]{0.95\textwidth}
    \centering
    \includegraphics[width=0.32\textwidth]{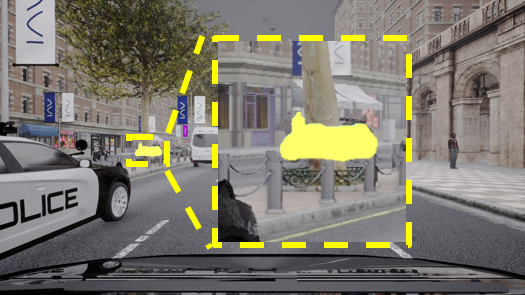}
        \includegraphics[width=0.32\textwidth]{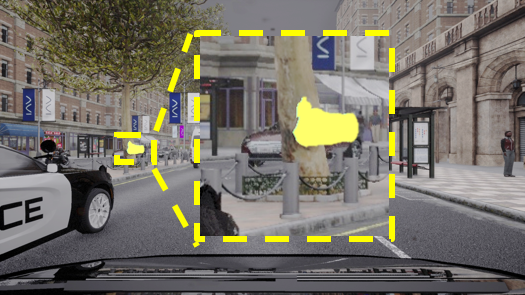}
            \includegraphics[width=0.32\textwidth]{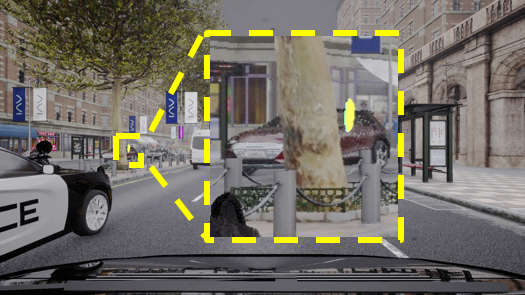}
        \includegraphics[width=0.32\textwidth]{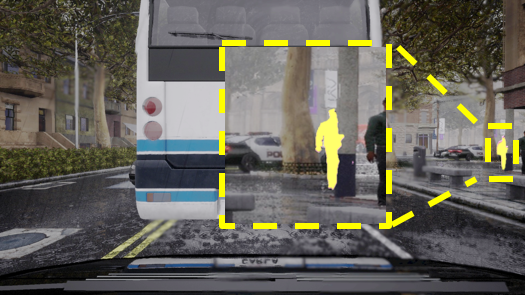}
            \includegraphics[width=0.32\textwidth]{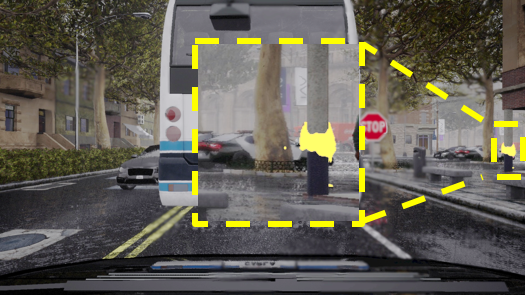}
                \includegraphics[width=0.32\textwidth]{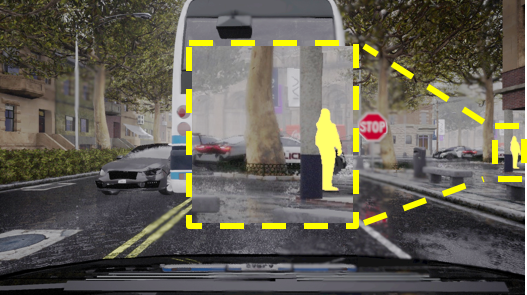}
                    \includegraphics[width=0.32\textwidth]{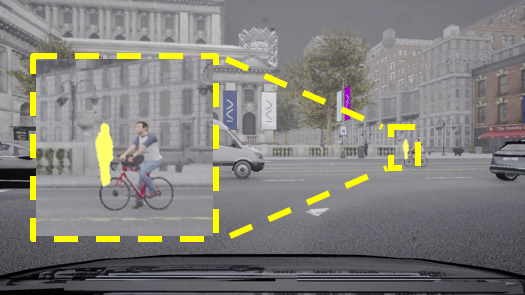}
        \includegraphics[width=0.32\textwidth]{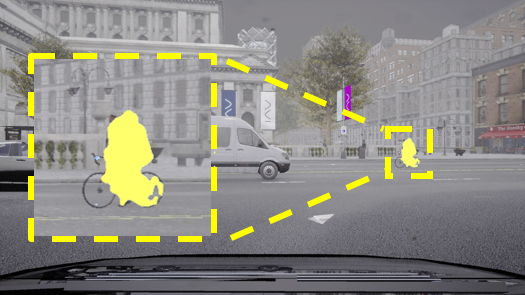}
            \includegraphics[width=0.32\textwidth]{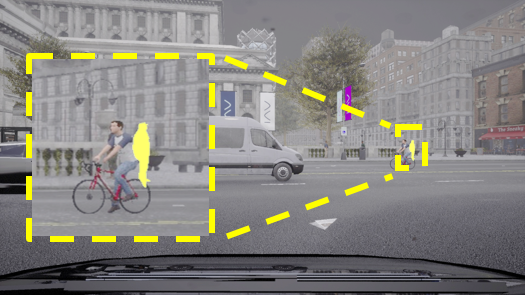}
                \end{minipage}
\begin{minipage}[t]{0.95\textwidth}
    \centering
$t-1$ \; \; \; \;\; \; \;\; \; \; \; \;\; \; \; \; \;\; \  $t$ \ \; \; \;\; \; \;\;\; \; \; \; \; \; \; \; \; \; \; $t+1$
\end{minipage}
\vspace{-5pt}
                \caption{\textbf{Failure Cases} of the proposed \texttt{S-AModal} method for three sequences $\mathbf{x}_{t-1}^{t+1}$ with overlayed colorized amodal predictions $\mathbf{a}_{t-1}^{t+1}$ on AmodalSynthDrive $\mathcal{D}^\text{val}_\text{ASD}$.}
    \label{fig:limitations-asd}
        \vspace{-18pt}
\end{figure}

We also investigate different point selection methods. For our main results in Tables \ref{tab:image-results-samadpt}, \ref{tab:video-results-asd-amodal}, we select points randomly from the visible mask. In Table \ref{tab:ablation-point-selection}, we show results on image level when instead of selecting a random point, we select the point with the highest saliency of the visible mask \cite{Dai2023}. Moreover, we show results when applying the erosion algorithm to the visible mask to ensure, we do not sample our point prompt from the boarder regions of an instance. We report results for erosion using two different kernel sizes: the default size of $3 \times 3$ (default \cite{VanderWalt2014}) and the size $7 \times 7$, which led to the best performance (best). As can be seen by the results in Table \ref{tab:ablation-point-selection}, sampling the point with the highest saliency does not lead to better results since it does not provide more information to amodal \texttt{SAM}. However, ensuring through the erosion algorithm \cite{VanderWalt2014} that we do not sample point prompts from boarder regions does lead to impressive performance gains in all metrics, e.g., an AP performance improvement of $1.12\%$ absolute from random sampling to the sampling after default erosion ($48.03\%$), and even $2.30\%$ absolute from random sampling to sampling after erosion using the kernel size $7 \times 7$ ($49.21\%$). Note that this simple post-processing of the visible segmentation masks leads to significant performance improvements while only adding a small computational overhead, i.e., the application of erosion per visible mask. The results in Table \ref{tab:ablation-point-selection} support our hypothesis that point prompts sampled from ambiguous boarder regions of an instance confuse the amodal \texttt{SAM} method, resulting in failure cases as shown in Figure \ref{fig:limitations-asd}. This highlights the potential of prompt engineering \cite{Shtedritski2023,Gu2023,Dai2023,Luddecke2022,Radford2021}  for this task and opens up a new research direction to design powerful prompts for amodal segmentation.


\begin{figure}[t]
    \centering
        \begin{minipage}[t]{0.02\textwidth}
        \rotatebox{90}{\hspace{10pt} Video 1}
        \rotatebox{90}{\hspace*{-1.4cm} Video 2}
        \rotatebox{90}{\hspace*{-3.1cm} Video 3}
        \rotatebox{90}{\hspace*{-4.8cm} Video 4}
        \rotatebox{90}{\hspace*{-6.6cm} Video 5}
    \end{minipage}
\begin{minipage}[t]{0.95\textwidth}
    \centering
    \includegraphics[width=0.32\textwidth]{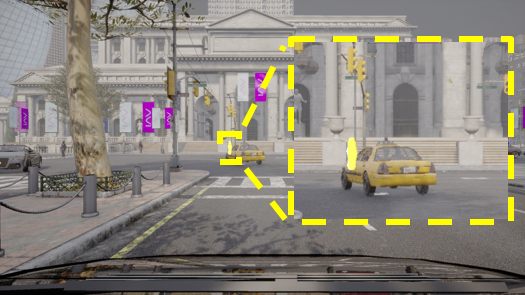}
        \includegraphics[width=0.32\textwidth]{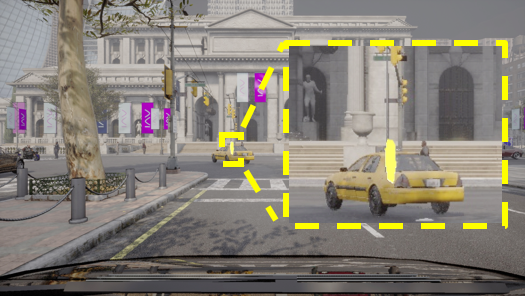}
            \includegraphics[width=0.32\textwidth]{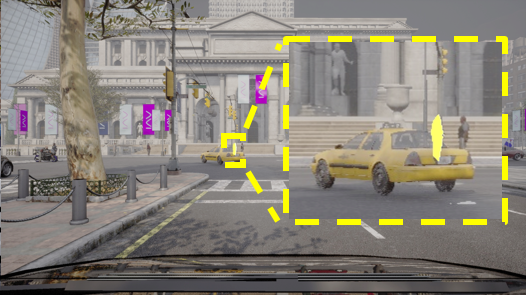}
    \includegraphics[width=0.32\textwidth]{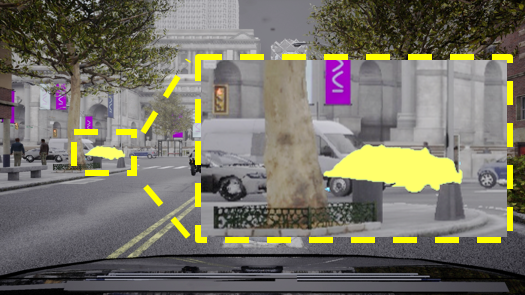}
        \includegraphics[width=0.32\textwidth]{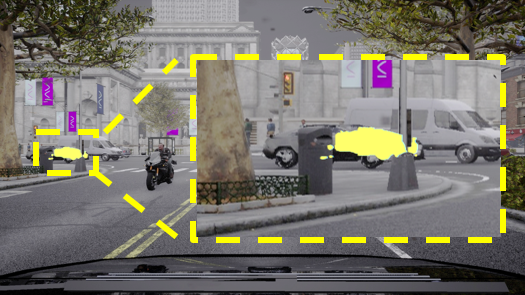}
            \includegraphics[width=0.32\textwidth]{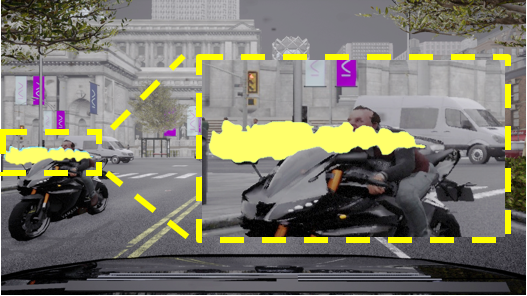}
                \includegraphics[width=0.32\textwidth]{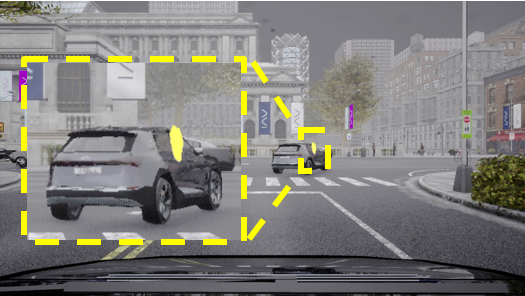}
        \includegraphics[width=0.32\textwidth]{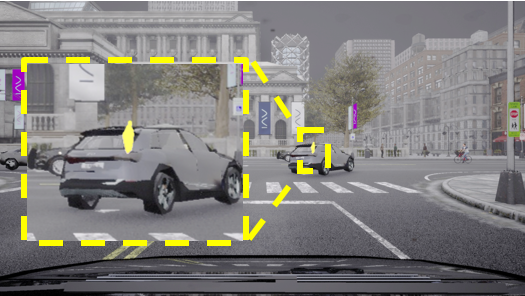}
            \includegraphics[width=0.32\textwidth]{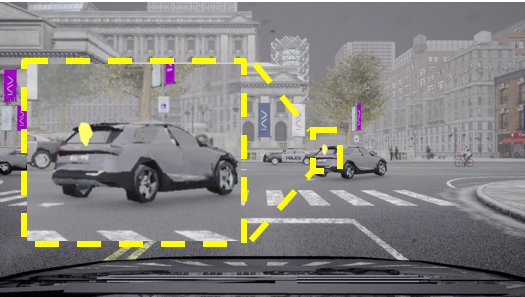}
                            \includegraphics[width=0.32\textwidth]{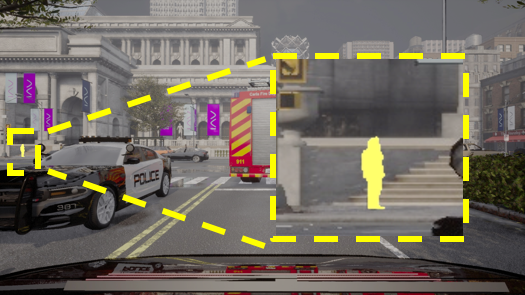}
        \includegraphics[width=0.32\textwidth]{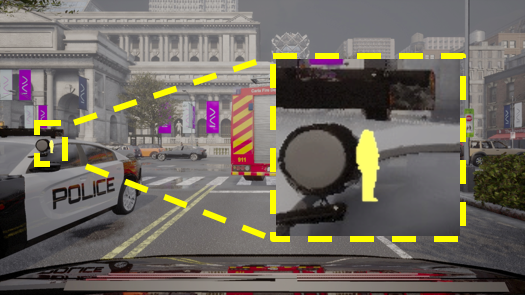}
            \includegraphics[width=0.32\textwidth]{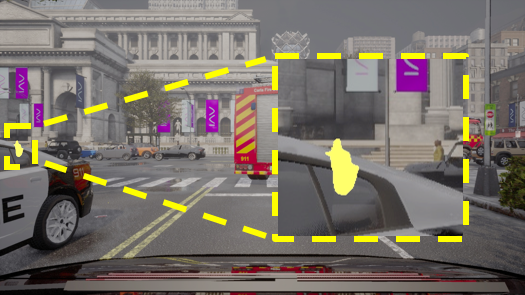}
                                        \includegraphics[width=0.32\textwidth]{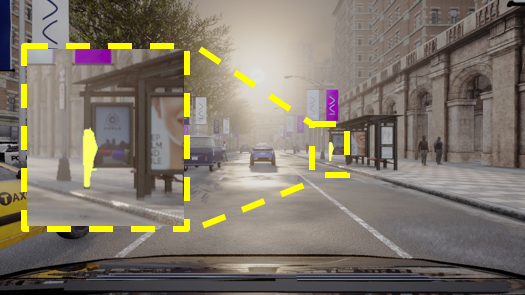}
        \includegraphics[width=0.32\textwidth]{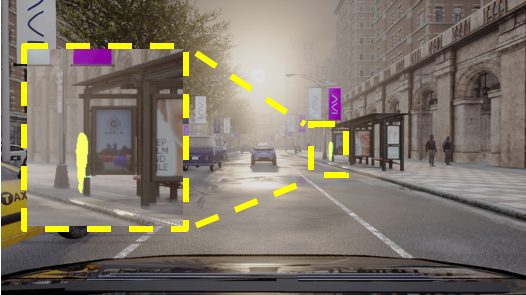}
            \includegraphics[width=0.32\textwidth]{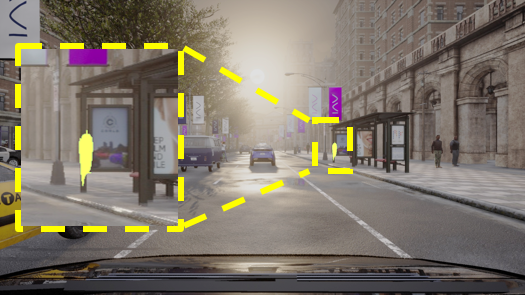}
                \end{minipage}
\begin{minipage}[t]{0.95\textwidth}
    \centering
$t-1$ \; \; \; \;\; \; \;\; \; \; \; \;\; \; \; \; \;\; \  $t$ \ \; \; \;\; \; \;\;\; \; \; \; \; \; \; \; \; \; \; $t+1$
\end{minipage}
\vspace{-7pt}
                \caption{\textbf{Qualitative results} of the proposed \texttt{S-AModal} method for five sequences $\mathbf{x}_{t-1}^{t+1}$ with overlayed colorized amodal predictions $\mathbf{a}_{t-1}^{t+1}$ on AmodalSynthDrive $\mathcal{D}^\text{val}_\text{ASD}$.}
    \label{fig:additional-good-examples-asd}
        \vspace{-15pt}
\end{figure}

\noindent \textbf{Additional qualitative results}:
Figure \ref{fig:additional-good-examples-asd} shows additional qualitative results of our proposed \texttt{S-AModal} on three videos of $\mathcal{D}^\text{val}_\text{ASD}$. In Video 1, a person is heavily occluded by the turning taxi. \texttt{S-AModal} recovers the shape reliably throughout this occlusion.
Video 2 shows in this case an occluded car. It vanishes behind a white truck in frame $t-1$. In the middle frame $t$ the car is completely occluded but its position and shape are reasonably recovered by \texttt{S-AModal}. The reappearance of the car in frame $t+1$ is also challenging to segment amodally in this case due to multi-layer occlusion of the car by the white van, another car and the motorcyclist in front. In this complex scenario, \texttt{S-AModal} predicts a slightly too large amodal mask for the car in frame $t+1$. 
In Video 3, a pedestrian vanishes behind a turning car. In frame $t-1$, just part of the head is still visible. The person is completely occluded in frame $t$ ans reappears in frame $t+1$. We see that in both partial occlusions ($t-1,t+1$) only a small part of the person is visible, and still \texttt{S-AModal} recovers the full shape of the person. In frame $t$, by relying on point tracking, the full shape of the person is predicted behind the car.
Video 4 shows again a pedestrian fully occluded by a bypassing car. Also in this case, \texttt{S-AModal} is able to recover the full occlusion in frame $t$.
In Video 5, a pedestrian is heavily occluded by a bus stop. In frames $t-1, t+1$, the feet are partly visible and hence, we are able to prompt the amodal \texttt{SAM} method to recover the full shape. In frame $t$, the pedestrian is not visible. Hence in this case, point tracking allows us to predict the amodal mask in frame $t$.

\noindent \textbf{Limitations}: Figure \ref{fig:limitations-asd} illustrates limitations of our method. In Video 1, a person walks behind a tree and reappears. However, in frame $t-1$, where the person is partially visible, amodal \texttt{SAM} predicts a wrong mask due to similar textures of person and car, which is then propagated to predict the amodal instance mask in the total occlusion in frame $t$.
In Video 2, a person is occluded by a pillar. While the shape of the person is recovered in frames $t-1$ and $t+1$, the more challenging occlusion in frame $t$ cannot be fully resolved: Only the upper body is predicted. We attribute this mainly to two reasons: First, the occlusion in frame $t$, where object parts are visible to the left and right of an occluder, is more often seen for the typically wider vehicle classes whose full shape is similar to the one recovered in frame $t$. Second, Video 2 is affected by challenging conditions like heavy rain and low light.
In Video 3, a pedestrian in frame $t$ is heavily but not fully occluded by a bypassing cyclist. In frames $t-1$ and $t+1$, the pedestrian is correctly segmented, however in frame $t$, only a very small part of the pedestrian is visible behind the cyclist and \texttt{S-AModal} is prompted using a point close to the cyclist. This leads to a confusion of the instances and the faulty segmentation of the cyclist in frame $t$. Future work could address these limitations by developing stronger prompts for amodal segmentation and adapting the newly published \texttt{SAM 2} \cite{Ravi2024} to this task to enhance prediction consistency.


    \vspace{-3pt}
\section{Conclusions}
\label{sec:conclusion}
    \vspace{-2pt}
In this work, we propose \texttt{S-AModal} for amodal video instance segmentation (VIS) with a focus on automated driving. To our knowledge, this is the first work to incorporate foundation models into the amodal segmentation task. We show that it is possible to adapt the original \texttt{SAM} model to the prompted amodal instance segmentation task using point prompts. Moreover, we show that incorporating this model into a VIS pipeline leads to an amodal VIS method with state-of-the-art (SOTA) performance while not relying on video-wise amodal VIS labels. Additionally, by incorporating the recent advances in point tracking into our pipeline, we are able to surpass limitations of previous amodal segmentation methods on both image- and video-level metrics and are able to provide instance masks also for temporal full occlusions of instances in the video sequence. 

\clearpage  

%
%
\bibliographystyle{splncs04}
\bibliography{main}

\begin{thebibliography}{10}
\providecommand{\url}[1]{\texttt{#1}}
\providecommand{\urlprefix}{URL }
\providecommand{\doi}[1]{https://doi.org/#1}

\bibitem{Ao2023}
Ao, J., Ke, Q., Ehinger, K.A.: {Image Amodal Completion: A Survey}. Computer Vision and Image Understanding  \textbf{229}, pp. 1--18 (Mar 2023)

\bibitem{Ao2024}
Ao, J., Ke, Q., Ehinger, K.A.: {Amodal Intra-class Instance Segmentation: Synthetic Datasets and Benchmark}. In: Proc. of WACV. pp. 281--290. Waikoloa, HI, USA (Jan 2024)

\bibitem{Back2022}
Back, S., Lee, J., Kim, T., Noh, S., Kang, R., Bak, S., Lee, K.: {Unseen Object Amodal Instance Segmentation via Hierarchical Occlusion Modeling}. In: Proc. of ICRA. pp. 5085--5092. Philadelphia, PA, USA (May 2022)

\bibitem{Breitenstein2022}
Breitenstein, J., Fingscheidt, T.: {Amodal Cityscapes: A New Dataset, its Generation, and an Amodal Semantic Segmentation Challenge Baseline}. In: Proc. of IV. pp. 1018--1025. Aachen, Germany (Jun 2022)

\bibitem{Breitenstein2023}
Breitenstein, J., Jin, K., Hakiri, A., Klingner, M., Fing\-scheidt, T.: {End-to-end Amodal Video Instance Segmentation}. In: Proc. of BMVC - Workshops. pp. 1--15. Aberdeen, UK (Nov 2023)

\bibitem{Breitenstein2022a}
Breitenstein, J., L\"ohdefink, J., Fingscheidt, T.: {Joint Prediction of Amodal and Visible Semantic Segmentation for Automated Driving}. In: Proc. of ECCV - Workshops. pp. 633--645. Tel Aviv, Israel (Oct 2022)

\bibitem{Chen2023}
Chen, T., Zhu, L., Ding, C., Cao, R., Zhang, S., Wang, Y., Li, Z., Sun, L., Mao, P., Zang, Y.: {SAM-Adapter: Adapting Segment Anything in Underperformed Scenes}. In: Proc. of ICCV - Workshops. pp. 3367--3375. Paris, France (Oct 2023)

\bibitem{Cheng2023}
Cheng, Y., Li, L., Xu, Y., Li, X., Yang, Z., Wang, W., Yang, Y.: {Segment and Track Anything}. arXiv  \textbf{2305.06558}, pp.~1--8 (May 2023)

\bibitem{Cordts2016}
Cordts, M., Omran, M., Ramos, S., Rehfeld, T., Enzweiler, M., Benenson, R., Franke, U., Roth, S., Schiele, B.: {The Cityscapes Dataset for Semantic Urban Scene Understanding}. In: Proc. of CVPR. pp. 3213--3223. Las Vegas, NV, USA (Jun 2016)

\bibitem{Dai2023}
Dai, H., Ma, C., Yan, Z., Liu, Z., Shi, E., Li, Y., Shu, P., Wei, X., Zhao, L., Wu, Z., Zeng, F., Zhu, D., Liu, W., Li, Q., Sun, L., Liu, S.Z.T., Li, X.: {SAMAug: Point Prompt Augmentation for Segment Anything Model}. arXiv  \textbf{2307.01187}, pp. 1--16 (Jul 2023)

\bibitem{Doersch2022}
Doersch, C., Gupta, A., Markeeva, L., Recasens, A., Smaira, L., Aytar, Y., Carreira, J., Zisserman, A., Yang, Y.: {TAP-Vid: A Benchmark for Tracking Any Point in a Video}. In: Proc. of NeurIPS. pp. 13610--13626. {New Orleans, LA, USA} (Dec 2022)

\bibitem{Doersch2023}
Doersch, C., Yang, Y., Vecerik, M., Gokay, D., Gupta, A., Aytar, Y., Carreira, J., Zisserman, A.: {TAPIR: Tracking Any Point with per-Frame Initialization and Temporal Refinement}. In: Proc. of ICCV. pp. 10061--10072. Paris, France (Oct 2023)

\bibitem{Dosovitskiy2021}
Dosovitskiy, A., Beyer, L., Kolesnikov, A., Weissenborn, D., Zhai, X., Unterthiner, T., Dehghani, M., Minderer, M., Heigold, G., Gelly, S., Uszkoreit, J., Houlsby, N.: {An Image is Worth 16x16 Words: Transformers for Image Recognition at Scale}. In: Proc.\ of {ICLR}. pp. 1--21. virtual (May 2021)

\bibitem{Dosovitskiy2017}
Dosovitskiy, A., Ros, G., Codevilla, F., Lopez, A., Koltun, V.: {CARLA: An Open Urban Driving Simulator}. In: Proc. of CoRL. pp. 1--16. Mountain View, CA, USA (Nov 2017)

\bibitem{fingscheidt_dnndataautomateddriving}
Fingscheidt, T., Gottschalk, H., Houben, S. (eds.): {Deep Neural Networks and Data for Automated Driving: Robustness, Uncertainty Quantification, and Insights Towards Safety}. {S}pringer {N}ature, Cham (2022). \doi{10.1007/978-3-031-01233-4}, \url{https://library.oapen.org/handle/20.500.12657/57375}

\bibitem{Follmann2019}
Follmann, P., K\"onig, R., H\"artinger, P., Klostermann, M.: {Learning to See the Invisible: End-to-End Trainable Amodal Instance Segmentation}. In: Proc. of WACV. pp. 1328--1336. Waikoloa Village, HI, USA (Jan 2019)

\bibitem{Geiger2013}
Geiger, A., Lenz, P., Stiller, C., Urtasun, R.: {Vision Meets Robotics: The KITTI Dataset}. International Journal of Robotics Research (IJRR)  \textbf{32}(11), pp. 1231--1237 (Aug 2013)

\bibitem{Gu2023}
Gu, J., Han, Z., Chen, S., Beirami, A., He, B., Zhang, G., Liao, R., Qin, Y., Tresp, V., Torr, P.: {A Systematic Survey of Prompt Engineering on Vision-Language Foundation Models}. arXiv  \textbf{2307.12980}, pp. 1--21 (Jul 2023)

\bibitem{Harley2022}
Harley, A.W., Fang, Z., Fragkiadaki, K.: {Particle Video Revisited: Tracking Through Occlusions Using Point Trajectories}. In: Proc. of ECCV. pp. 59--75. Tel Aviv, Israel (Oct 2022)

\bibitem{Heo2023}
Heo, M., Hwang, S., Hyun, J., Kim, H., Oh, S.W., Lee, J.Y., Kim, S.J.: {A Generalized Framework for Video Instance Segmentation}. In: Proc. of CVPR. pp. 14623--14632. {Vancouver, BC, Canada} (Jun 2023)

\bibitem{Heo2022}
Heo, M., Hwang, S., Oh, S.W., Lee, J.Y., Kim, S.J.: {VITA: Video Instance Segmentation via Object Token Association}. In: Proc. of NeurIPS. pp. 1--12. New Orleans, LA, USA (Dec 2022)

\bibitem{Hu2019a}
Hu, Y.T., Chen, H.S., Hui, K., Huang, J.B., Schwing, A.G.: {SAIL-VOS: Semantic Amodal Instance Level Video Object Segmentation -- A Synthetic Dataset and Baselines}. In: Proc. of CVPR. pp. 3105--3115. Long Beach, CA, USA (Jun 2019)

\bibitem{Karaev2023}
Karaev, N., Rocco, I., Graham, B., Neverova, N., Vedaldi, A., Rupprecht, C.: {CoTracker: It is Better to Track Together}. arXiv  \textbf{2307.07635}, pp. 1--13 (Jul 2023)

\bibitem{Ke2021}
Ke, L., Tai, Y.W., Tang, C.K.: {Deep Occlusion-Aware Instance Segmentation with Overlapping BiLayers}. In: Proc. of CVPR. pp. 4019--4028. Nashville, TN, USA (Jun 2021)

\bibitem{Ke2023}
Ke, L., Ye, M., Danelljan, M., Liu, Y., Tai, Y.W., Tang, C.K., Yu, F.: {Segment Anything in High Quality}. In: Proc. of NeurIPS. pp. 29914--29934. Vancouver, BC, Canada (Dec 2023)

\bibitem{Kirillov2023}
Kirillov, A., Mintun, E., Ravi, N., Mao, H., Rolland, C., Gustafson, L., Xiao, T., Whitehead, S., Berg, A.C., Lo, W.Y., Doll{\'a}r, P., Girshick, R.: {Segment Anything}. In: Proc. of ICCV. pp. 4015--4026. Paris, France (Oct 2023)

\bibitem{Li2016a}
Li, K., Malik, J.: {Amodal Instance Segmentation}. In: {Proc. of ECCV}. pp. 677--693. Amsterdam, The Netherlands (Oct 2016)

\bibitem{Liao2021}
Liao, Y., Xie, J., Geiger, A.: {KITTI-360: A Novel Dataset and Benchmarks for Urban Scene Understanding in 2D and 3D}. arXiv  \textbf{2109.13410}, pp. 1--32 (Sep 2021)

\bibitem{Lin2014microsoft}
Lin, T.Y., Maire, M., Belongie, S., Hays, J., Perona, P., Ramanan, D., Doll{\'a}r, P., Zitnick, C.L.: {Microsoft COCO: Common Objects in Context}. In: Proc. of ECCV. pp. 740--755. Zurich, Switzerland (Sep 2014)

\bibitem{Ling2020}
Ling, H., Acuna, D., Kreis, K., Kim, S.W., Fidler, S.: {Variational Amodal Object Completion}. In: Proc. of NeurIPS. pp. 16246--16257. Vancouver, BC, Canada (Dec 2020)

\bibitem{Loshchilov2019}
Loshchilov, I., Hutter, F.: {Decoupled Weight Decay Regularization}. In: Proc. of ICLR. pp. 1--18. New Orleans, LA, USA (May 2019)

\bibitem{Luddecke2022}
L{\"u}ddecke, T., Ecker, A.: {Image Segmentation Using Text and Image Prompts}. In: Proc. of CVPR. pp. 7086--7096. New Orleans, LA, USA (Jun 2022)

\bibitem{Mohan2022}
Mohan, R., Valada, A.: {Amodal Panoptic Segmentation}. In: Proc. of CVPR. pp. 21023--21032. New Orleans, LA, USA (Jun 2022)

\bibitem{Nguyen2021}
Nguyen, K., Todorovic, S.: {A Weakly Supervised Amodal Segmenter with Boundary Uncertainty Estimation}. In: Proc. of ICCV. pp. 2995--3003. Virtual (Oct 2021)

\bibitem{Purkait2019}
Purkait, P., Zach, C., Reid, I.D.: {Seeing Behind Things: Extending Semantic Segmentation to Occluded Regions}. In: Proc. of IROS. pp. 1998--2005. Macau, SAR, China (Nov 2019)

\bibitem{Qi2019}
Qi, L., Jiang, L., Liu, S., Shen, X., Jia, J.: {Amodal Instance Segmentation With KINS Dataset}. In: Proc. of CVPR. pp. 3014--3023. Long Beach, CA, USA (Jun 2019)

\bibitem{Radford2021}
Radford, A., Kim, J.W., Hallacy, C., Ramesh, A., Goh, G., Agarwal, S., Sastry, G., Askell, A., Mishkin, P., Clark, J., Krueger, G., Sutskever, I.: {Learning Transferable Visual Models From Natural Language Supervision}. In: Proc. of ICML. pp. 8748--8763. virtual (Jul 2021)

\bibitem{Rajic2023}
Rajič, F., Ke, L., Tai, Y.W., Tang, C.K., Danelljan, M., Yu, F.: {Segment Anything Meets Point Tracking}. arXiv  \textbf{2307.01197}, pp. 1--15 (Dec 2023)

\bibitem{Ravi2024}
Ravi, N., Gabeur, V., Hu, Y.T., Hu, R., Ryali, C., Ma, T., Khedr, H., R{\"a}dle, R., Rolland, C., Gustafson, L., Mintun, E., Pan, J., Alwala, K.V., Carion, N., Wu, C.Y., Girshick, R., Doll{\'a}r, P., Feichtenhofer, C.: {SAM 2: Segment Anything in Images and Videos}. arXiv  \textbf{2408.00714}, pp. 1--41 (Aug 2024)

\bibitem{Reddy2022}
Reddy, N.D., Tamburo, R., Narasimhan, S.: {WALT: Watch And Learn 2D Amodal Representation using Time-lapse Imagery}. In: Proc. of CVPR. pp. 9356--9366. New Orleans, LA, USA (Jun 2022)

\bibitem{Sekkat2023}
Sekkat, A.R., Mohan, R., Sawade, O., Matthes, E., Valada, A.: {AmodalSynthDrive: A Synthetic Amodal Perception Dataset for Autonomous Driving}. arXiv  \textbf{2309.06547}, pp. 1--12 (Sep 2023)

\bibitem{Shaharabany2023}
Shaharabany, T., Dahan, A., Giryes, R., Wolf, L.: {AutoSAM: Adapting SAM to Medical Images by Overloading the Prompt Encoder}. In: Proc. of BMVC. pp. 1--15. Aberdeen, UK (Nov 2023)

\bibitem{Shtedritski2023}
Shtedritski, A., Rupprecht, C., Vedaldi, A.: {What Does CLIP Know About a Red Circle? Visual Prompt Engineering for VLMs}. In: Proc. of ICCV. pp. 11987--11997. Paris, France (Oct 2023)

\bibitem{Sun2022}
Sun, Y., Kortylewski, A., Yuille, A.: {Amodal Segmentation through Out-of-Task and Out-of-Distribution Generalization with a Bayesian Model}. In: Proc. of CVPR. pp. 1215--1224. New Orleans, LA, USA (Jun 2022)

\bibitem{VanderWalt2014}
Van~der Walt, S., Sch{\"o}nberger, J.L., Nunez-Iglesias, J., Boulogne, F., Warner, J.D., Yager, N., Gouillart, E., Yu, T.: {scikit-image: Image Processing in Python}. PeerJ  \textbf{2}, p.~e453 (May 2014)

\bibitem{Wang2023}
Wang, Q., Chang, Y.Y., Cai, R., Li, Z., Hariharan, B., Holynski, A., Snavely, N.: {Tracking Everything Everywhere All at Once}. In: Proc. of ICCV. pp. 19795--19806. Paris, France (Oct 2023)

\bibitem{Wu2023}
Wu, J., Ji, W., Liu, Y., Fu, H., Xu, M., Xu, Y., Jin, Y.: {Medical SAM Adapter: Adapting Segment Anything Model for Medical Image Segmentation}. arXiv  \textbf{2304.12620}, pp. 1--10 (Apr 2023)

\bibitem{Xiao2024}
Xiao, A., Xuan, W., Qi, H., Xing, Y., Ren, R., Zhang, X., Ling, S., Lu, S.: {CAT-SAM: Conditional Tuning Network for Few-Shot Adaptation of Segmentation Anything Model}. arXiv  \textbf{2402.03631}, pp. 1--25 (Feb 2024)

\bibitem{Xie2024}
Xie, Z., Guan, B., Jiang, W., Yi, M., Ding, Y., Lu, H., Zhang, L.: {PA-SAM: Prompt Adapter SAM for High-quality Image Segmentation}. In: Proc. of ICME. pp. 1--10. Niagra Falls, Canada (Jul 2024)

\bibitem{Xu2020}
Xu, Z., Zhang, W., Tan, X., Yang, W., Huang, H., Wen, S., Ding, E., Huang, L.: {Segment as Points for Efficient Online Multi-Object Tracking and Segmentation}. In: Proc. of ECCV. pp. 1--17. Glasgow, UK (Aug 2020)

\bibitem{Yang2023}
Yang, J., Gao, M., Li, Z., Gao, S., Wang, F., Zheng, F.: {Track Anything: Segment Anything Meets Videos}. arXiv  \textbf{2304.11968}, pp.~1--7 (Apr 2023)

\bibitem{Yang2019c}
Yang, L., Fan, Y., Xu, N.: {Video Instance Segmentation}. In: Proc. of ICCV. pp. 5188--5197. Seoul, Korea (Oct 2019)

\bibitem{Yao2022}
Yao, J., Hong, Y., Wang, C., Xiao, T., He, T., Locatello, F., Wipf, D., Fu, Y., Zhang, Z.: Self-supervised {{Amodal Video Object Segmentation}}. In: Proc. of NeurIPS. pp. 6278--6291. {New Orleans, LA, USA} (Dec 2022)

\bibitem{Zheng2023}
Zheng, Y., Harley, A.W., Shen, B., Wetzstein, G., Guibas, L.J.: {PointOdyssey: A Large-Scale Synthetic Dataset for Long-Term Point Tracking}. In: Proc. of ICCV. pp. 19855--19865. Paris, France (Oct 2023)

\bibitem{Zhu2017a}
Zhu, Y., Tian, Y., Metaxas, D., Doll\'ar, P.: {Semantic Amodal Segmentation}. In: Proc. of CVPR. pp. 1464--1472. Honolulu, HI, USA (Jul 2017)

\end{thebibliography}

\newpage
\title{Supplementary:\\ Foundation Models for Amodal Video Instance Segmentation in Automated Driving} 

\titlerunning{Supplementary: Foundation Models for Amodal VIS in Automated Driving}

\author{Jasmin Breitenstein \inst{1} \and
Franz J\"unger \inst{1} \and Andreas B\"ar \inst{1} \and
Tim Fingscheidt \inst{1}}

\authorrunning{J. Breitenstein et al.}

\institute{$^1$Institute for Communications Technology \\
Technische Universit\"at Braunschweig \\
\email{\{j.breitenstein, f.juenger, andreas.baer, t.fingscheidt\}@tu-bs.de}}

\maketitle
\setcounter{section}{0}
\renewcommand{\thesection}{\Alph{section}} 

\noindent In this supplementary, we give additional information about our proposed \texttt{S-AModal} method for amodal video instance segmentation (amodal VIS).

\section{Amodal \texttt{SAM}}
\label{suppsec:amodalsam}

In this section we describe the adapter method for our amodal \texttt{SAM} as well as our chosen training strategy. Finally, we describe how our sampled point prompts are used as input to amodal \texttt{SAM}. 


\begin{figure}[t]
    \centering
        \includegraphics[width=\linewidth]{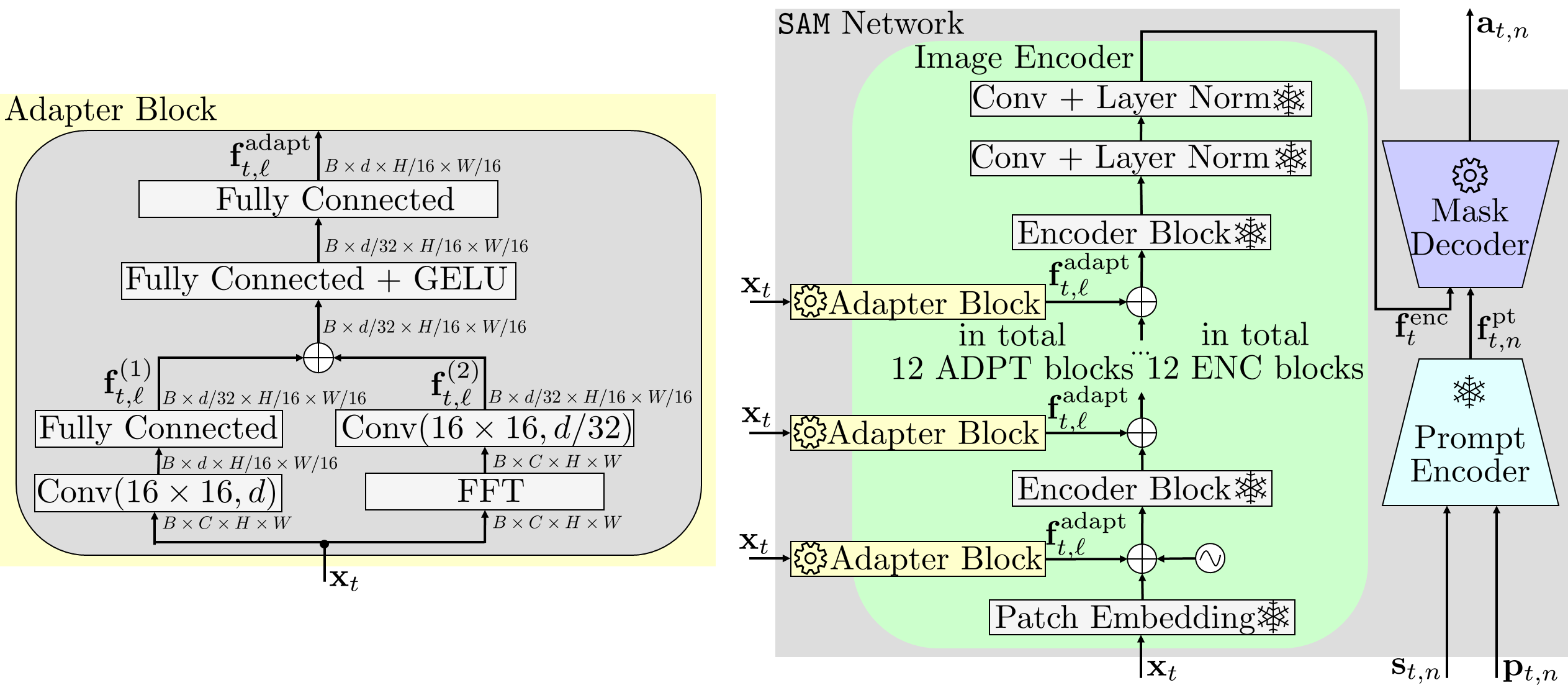}
                \caption{Detailed structure of the adapter block \cite{Chen2023} (left) and the \texttt{SAM} network \cite{Kirillov2023} used during our amodal fine-tuning (right). Snowflakes indicate layers frozen during fine-tuning while the gear wheel indicates adjustable layers. The fine-tuned \texttt{SAM} network is used in our \texttt{S-AModal} method as amodal \texttt{SAM} $\mathbf{f}^\text{aSAM}()$ network, as shown in Figure 2.}
    \label{fig:samnetwork}
\end{figure}

Figure \ref{fig:samnetwork} shows the network structure. Following the work of Chen et al.\ \cite{Chen2023}, we add adapter blocks to the image encoder. An adapter block follows each encoder block as can bee seen in Figure \ref{fig:samnetwork}. Internally, each adapter block extracts two types of features $\mathbf{f}^{(1)}_t, \mathbf{f}^{(2)}_t$, from the input image $\mathbf{x}_t$. After adding these features, each adapter block applies a fully connected layer with GELU activation. A second fully connected layer follows, which is shared between all adapter blocks to obtain the adapter features $\mathbf{f}^\text{adapt}_{t,\ell}$. The detailed structure is shown in Figure \ref{fig:samnetwork} (left). Note that all adapter block features are of course also dependent on the current adapter block in layer $\ell \in \lbrace 1,...,12\rbrace$.
On the right of Figure \ref{fig:samnetwork}, the overall network architecture of the \texttt{SAM} model \cite{Kirillov2023} is shown in the form as we use it for amodal fine-tuning: In the image encoder, the adapter blocks are introduced between the encoder blocks. The output features of each adapter block $\mathbf{f}^\text{adapt}_{t,\ell}$ are added to the features of the image encoder before each encoder block. The final encoder features $\mathbf{f}^\text{enc}_t$ are one of the inputs to the mask decoder. The second input to the mask decoder are the features obtained from the prompt encoder $\mathbf{f}^\text{pt}_{t,n}$. The prompt encoder receives as input a point K-tuple $\mathbf{p}_{t,n}= \left( p_{t,n,k}\right) \in \lbrace1,..,H\cdot W\rbrace^K$ where $n \in \mathcal{N}_t = \lbrace 1, \ldots, N_t \rbrace$ denoting the instance index with $N_t$ being the number of instances observed until frame $t$ and $H,W$ being height and width, respectively. Additionally, the prompt encoder receives corresponding labels for the point K-tuple. We denote these labels as $\mathbf{s}_{t,n}=\left( s_{t,n,k} \right) \in \lbrace 0,1 \rbrace^{K}$. These labels describe whether each point prompt is positive ($s_{t,n,k}=1$) or negative ($s_{t,n,k}=0$), i.e., whether the point is part of the desired mask. Since we use point prompts derived from predicted instance masks, we only use positive point prompts with $s_{t,n,k}=1$.
Given both the final encoder features $\mathbf{f}^\text{enc}_t$ and the features $\mathbf{f}^\text{pt}_{t,n}$ of the prompt encoder, the mask decoder then outputs the amodal mask $\mathbf{a}_{t,n}$. 
In contrast to Chen et al.\ \cite{Chen2023}, we keep the prompt encoder for our network. Given the input image $\mathbf{x}_t$, we obtain encoder features $\mathbf{f}^\text{enc}_t$ from the image encoder, and given a point prompt $p_{t,n}$, we obtain point encoder features $\mathbf{f}^\text{pt}_{t,n}$ from the prompt encoder. Both features are then input to the mask decoder to obtain the final amodal mask $\mathbf{a}_{t,n}$. 
During the fine-tuning process, we keep the original image encoder blocks and the prompt encoder frozen. Only the mask decoder as well as the adapter blocks remain adjustable.
As loss function, we use the common choice of combining dice and focal loss \cite{Chen2023,Ke2023,Kirillov2023,Shaharabany2023,Wu2023}, and the AdamW optimizer \cite{Loshchilov2019}. We observe that for our purpose, training with small batch sizes ($B=1$) was more advantageous.

\begin{algorithm}[t]
\SetAlgoLined
\KwIn{visible mask $\mathbf{m}_{t,n}$, input image $\mathbf{x}_t$, number of desired point prompts $K$, amodal \texttt{SAM} model with image encoder $\mathbf{E}$, prompt encoder $\mathbf{P}$ and mask decoder $\mathbf{D}$}
\KwOut{amodal mask $\mathbf{a}_{t,n}$}
$\mathcal{I}^{(\mathbf{m})}_{t,n}$ = $\lbrace i \in \mathcal{I}=\lbrace 1, ..., H\cdot W \rbrace | \mathbf{m}_{t,n}(i)=1 \rbrace$ \\
$\mathbf{p}_{t,n}= \left(p_{t,n,k}\right)$ = random.choice($\mathcal{I}^{(\mathbf{m})}_{t,n}$,K)\\
$\mathbf{s}_{t,n} = \mathbbm{1}^K$\\
$\mathbf{f}^\text{enc}_t$ = $\mathbf{E}(\mathbf{x}_t)$\\
$\mathbf{f}^\text{pt}_{t,n}$ = $\mathbf{P}(\mathbf{p}_{t,n},\mathbf{s}_{t,n})$\\
$\mathbf{a}_{t,n}$ = $\mathbf{D}(\mathbf{f}^\text{enc}_t,\mathbf{f}^\text{pt}_{t,n}$)\\
\Return{$\mathbf{a}_{t,n}$}
\caption{Prompting amodal \texttt{SAM} with points from a visible mask}
\label{alg:pointprompts}
\end{algorithm}

\noindent \textbf{Point prompts for amodal \texttt{SAM}}: Here, we briefly describe how points are extracted from a given visible mask and used to prompt amodal \texttt{SAM} $\mathbf{f}^\text{aSAM}$. Note that in the absence of a visible mask, we apply point tracking to move a previously predicted amodal mask along the predicted trajectory instead of prompting amodal \texttt{SAM}. The algorithm to prompt amodal \texttt{SAM} is shown in Algorithm \ref{alg:pointprompts} in pseudo code. Given a visible mask $\mathbf{m}_{t,n}$, an input image $\mathbf{x}_t$, the desired number of point prompts $K$ and the amodal \texttt{SAM} model with image encoder $\mathbf{E}$, prompt encoder $\mathbf{P}$ and mask decoder $\mathbf{D}$, the output is the amodal mask $\mathbf{a}_{t,n}$. The visible mask $\mathbf{m}_{t,n} \in \lbrace 0,1 \rbrace^{H \times W}$ takes on the value $1$ for the pixel indices $i \in \mathcal{I}=\lbrace 1, ..., H \cdot W \rbrace$ where the mask is predicted, i.e.,\ $m_{t,n}(i)=1$. We extract the subset of pixel indices $\mathcal{I}^{(\mathbf{m})}_{t,n}$ where the visible mask is predicted. 
From the set $\mathcal{I}^{(\mathbf{m})}_{t,n}$ we randomly sample $K$ point prompts $\mathbf{p}_{t,n}=(p_{t,n,k})$. As shown in Figure \ref{fig:samnetwork} (right), the prompt encoder takes as input the point prompts and corresponding labels $\mathbf{s}_{t,n} \in \lbrace 0,1 \rbrace^K$ indicating whether each point is positive ($1$), i.e.\ belongs to the desired mask, or negative ($0$), i.e.\ does not belong to the desired mask. Due to our choice of sampling the point prompts $\mathbf{p}_{t,n}$ from the visible mask $\mathbf{m}_{t,n}$, we only consider positive point labels, i.e.,\ $\mathbf{s}_{t,n,k}=1$.
We obtain the encoder features from the image encoder, i.e.\ $\mathbf{E}(\mathbf{x}_t)$ = $\mathbf{f}^\text{enc}_t$, and the point prompt features from the prompt encoder, i.e., $\mathbf{P}(\mathbf{p}_{t,n},\mathbf{s}_{t,n})$ = $\mathbf{f}^\text{pt}_{t,n}$. Both are then input to the mask decoder to obtain the final amodal mask $\mathbf{a}_{t,n} = \mathbf{D}(\mathbf{f}^\text{enc}_t,\mathbf{f}^\text{pt}_{t,n})$.
For more details on the code, please refer to our github repository \url{https://github.com/ifnspaml/S-AModal}.

\begin{figure}[t]
    \centering
\begin{minipage}[t]{0.49\textwidth}
    \centering
    \includegraphics[width=\linewidth]{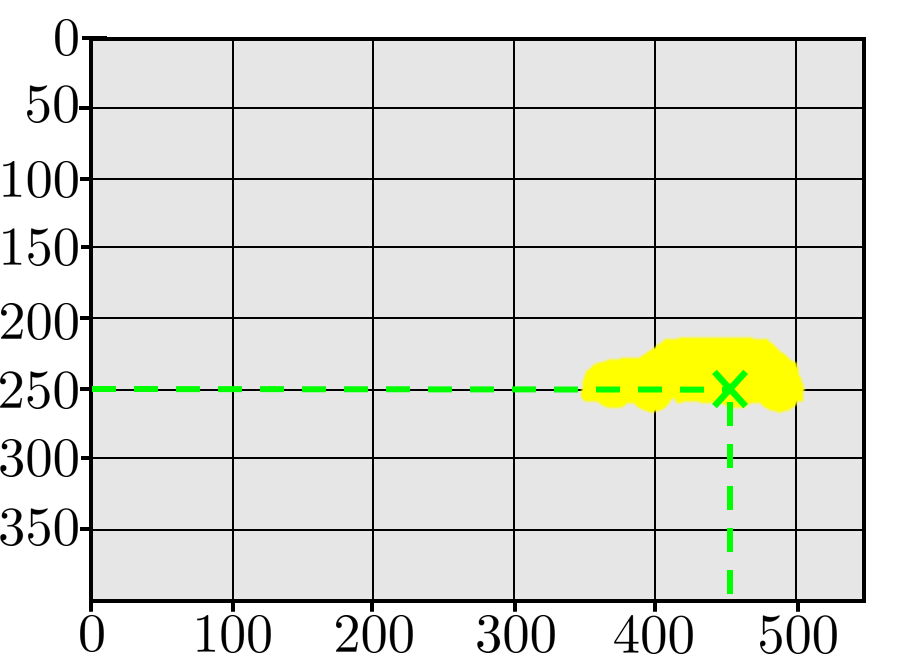}
    \end{minipage}
    \hfill
    \begin{minipage}[t]{0.49\textwidth}
        \includegraphics[width=\linewidth]{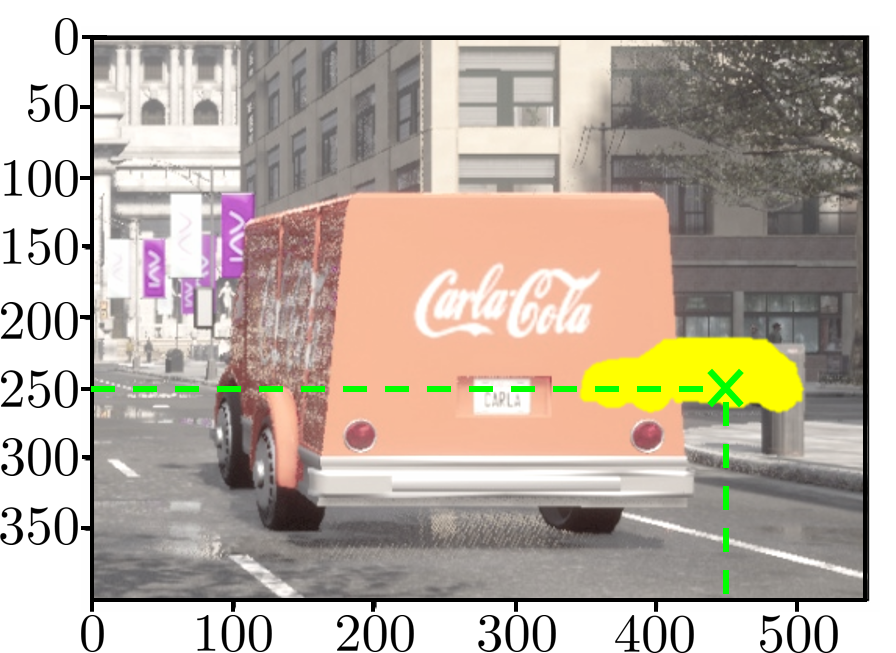}
                \end{minipage}
                \caption{Schematic visualization of a point prompt $p_{t,n,k}$ (left, green cross) resulting in the yellow amodal mask. For better understanding, we denote the point prompt using its height $h$ and width $w$ value, i.e., $(h,w)=(250,400)$ (indicated by green line). Right: visualization of the same point prompt (green cross) in the corresponding image of $\mathcal{D}_\text{ASD}^\text{val}$ again with height and width value $(h,w)=(250,450)$ (green lines) resulting in the yellow amodal mask.}
    \label{fig:pointprompt}
\end{figure}

For a better understanding, we visualize an example for a point prompt in Figure \ref{fig:pointprompt}. On the left, we show the amodal mask $\mathbf{a}_{t,n}$ of a car in yellow. The point prompt is indicated by a green cross with corresponding green lines to the width $w$ and height $h$ value corresponding to this point, i.e., $(h,w)=(250,450)$. In addition to the schematic figure, we show the corresponding image from $\mathcal{D}_\text{ASD}^\text{val}$ on the right side. The amodal mask $\mathbf{a}_{t,n}$ is shown yellow with the point prompt $p_{t,n}$ indicated again by the green cross with corresponding green lines to the height value $h=250$ and the width value $w=450$. Note that for this example, we visualized the point prompts for the choice $K=1$.  

\section{Metrics for amodal and visible VIS}
\label{suppsec:metrics}

Metrics for amodal VIS are derived from the metrics typically used in (visible) VIS. Here, one uses the notion of average precision (AP), i.e., the area under the precision-recall curve, as is typically used in instance segmentation. We follow the definition of MS-COCO \cite{Lin2014microsoft} to calculate AP at pre-defined intersection over union (IoU) thresholds, and AP at an IoU threshold of $0.5$ (AP$_{50}$).
However, AP relies on calculating true positive predictions by considering the IoU between the predicted instance mask $\mathbf{m}_{t,n}$ and the ground truth instance mask $\overline{\mathbf{m}}_{t,n}$. On videos, we need to slightly alter this notion, as we are not only interested in the image-wise segmentation quality, but also in a video-wise segmentation quality. Hence, we consider the overlap between predicted and ground truth instance masks over the entire video sequence \cite{Yang2019c},
\begin{equation}
    \text{vIoU}(\mathbf{m}_{1,n}^T,\overline{\mathbf{m}}_{1,n}^T) = \frac{\sum\limits_{t=1}^T |\mathbf{m}_{t,n} \cap \overline{\mathbf{m}}_{t,n}|}{\sum\limits_{t=1}^T |\mathbf{m}_{t,n} \cup \overline{\mathbf{m}}_{t,n}|},
    \label{eq:viou}
\end{equation}
where $\cap$ denotes the intersection between the predicted and ground-truth mask, $\cup$ means the union of both masks, and $|\cdot |$ is the cardinality. In this case, we define cardinality as the number of pixel indices $i$ where the union or intersection of both masks takes on the value $1$, i.e., the area of the resulting mask. From Equation \ref{eq:viou} it follows that a correct prediction needs to have sufficient overlap with all ground truth instance masks over a video sequence, and the tracking performance has a large influence on this video-wise IoU notion. To be able to better distinguish the video- and image-wise metrics, we use the notion vAP whenever performance on videos is addressed. 
Amodal metrics are calculated using the amodal masks $\mathbf{a}_{1,n}^T$. Next to standard AP, we also calculate metric variants as defined for the SAIL-VOS dataset and following the standard in literature \cite{Lin2014microsoft,Hu2019a,Breitenstein2023}, i.e., considering small, medium, and large objects, and partially and heavily occluded objects separately.

\section{Video Instance Segmentation Results on AmodalSynthDrive}
\label{suppsec:vis-results-asd}
To have a full understanding about our proposed method for amodal VIS, it is important to regard the (visible) VIS performance of the underlying VIS method. For our work, we apply the \texttt{GenVIS} method \cite{Heo2023}. \texttt{GenVIS} \cite{Heo2023} is a high-performing VIS method using a training strategy for sequential learning based on queries. Additionally, it introduces a memory to access information from previous states \cite{Heo2023}. 

\begin{table}[h]
    \centering
    \footnotesize
    \setlength{\tabcolsep}{4pt}
    {\renewcommand{\arraystretch}{1.0}
    \begin{tabular}{c|c|c|c|c|c|c|c|c}
    Method &
    Resolution &
    AP &
    AP$_\text{50}$ &
    AP$_\text{50}^\text{P}$ &
    AP$_\text{50}^\text{H}$ &
    AP$_\text{50}^\text{L}$ &
    AP$_\text{50}^\text{M}$ &
    AP$_\text{50}^\text{S}$ \\  
    \hline 
    \hline
    \texttt{VITA} \cite{Heo2022} & $540 \!\times\! 960$ & 16.45 & 28.27 & 34.64 & \phantom{0}9.30 & 63.12 & 51.55 & \phantom{0}\underline{5.71} \\
    \texttt{VITA} \cite{Heo2022}    & $1080 \!\times \! 1920$ & \underline{23.84} & \underline{33.58} & \underline{44.92} & \underline{14.31} & \underline{76.00} & \underline{57.95} & \phantom{0}5.50 \\
    \hline
    \texttt{GenVIS} \cite{Heo2023} & $540 \!\times \! 960$ & 22.73 & 32.84 & 44.48 & 11.74 & \textbf{85.53} & 53.50 & \phantom{0}2.58 \\
    \texttt{GenVIS} \cite{Heo2023} & $1080 \!\times \! 1920$ & \textbf{30.38} & \textbf{43.83} & \textbf{53.99} & \textbf{18.45} & 66.23 & \textbf{69.65} & \textbf{12.59} \\
    \end{tabular}}
    \caption{\textbf{Visible image}-level results by \texttt{VITA} \cite{Heo2022} and \texttt{GenVIS} \cite{Heo2023} on the AmodalSynthDrive validation dataset $\mathcal{D}^\text{val}_\text{ASD}$. Best results in \textbf{bold}, second best \underline{underlinded}.}
    \label{tab:asd_image}
\end{table}

Table \ref{tab:asd_image} shows results for \texttt{VITA} \cite{Heo2022} and its advanced version \texttt{GenVIS} \cite{Heo2023}. While \texttt{GenVIS} training takes longer than training \texttt{VITA}, Table \ref{tab:asd_image} clearly shows the advantage of the additional training time, where \texttt{GenVIS} reaches an AP of $30.38\%$, while \texttt{VITA} reaches only an AP of $23.84\%$. Additionally, we observe that using the original image resolution of $1080 \times 1920$ is much more advantageous compared to using half the resolution, as, e.g., AP$_{50}$ increases by $10.99\%$ absolute to $43.83\%$ for \texttt{GenVIS}.

\begin{table}[h]
    \centering
    \footnotesize
    \setlength{\tabcolsep}{4pt}
    {\renewcommand{\arraystretch}{1.0}
    \begin{tabular}{c|c|c|c|c|c|c|c|c}
    Method &
    Resolution &
    vAP &
    vAP$_\text{50}$ &
    vAP$_\text{50}^\text{P}$ &
    vAP$_\text{50}^\text{H}$ &
    vAP$_\text{50}^\text{L}$ &
    vAP$_\text{50}^\text{M}$ &
    vAP$_\text{50}^\text{S}$ \\  
    \hline 
    \hline
    \texttt{VITA} \cite{Heo2022} & $540 \!\times\! 960$ & 14.72 & \textbf{25.27} & \underline{36.12} & \textbf{27.10} & 47.92 & \underline{26.07} & \phantom{0}\textbf{4.57} \\
    \texttt{VITA} \cite{Heo2022}    & $1080 \!\times \! 1920$  & \underline{15.32} & \underline{24.88} & \textbf{46.07} & 19.41 & 48.75 & \textbf{48.32} & \phantom{0}\underline{3.37} \\
    \hline
    \texttt{GenVIS} \cite{Heo2023} & $540 \!\times \! 960$ & 14.46 & 21.49 & 34.03 & 21.00 & \textbf{63.79} & 24.89 & \phantom{0}0.59\\
    \texttt{GenVIS} \cite{Heo2023} & $1080 \!\times \! 1920$  & \textbf{16.36} & 23.13 & 35.47 & 21.81 & \underline{55.24} & 15.09 & \phantom{0}1.71 \\
    \end{tabular}}
    \caption{\textbf{Visible video}-level results by \texttt{VITA} \cite{Heo2022} and \texttt{GenVIS} \cite{Heo2023} on the AmodalSynthDrive validation dataset $\mathcal{D}^\text{val}_\text{ASD}$. Best results in \textbf{bold}, second best \underline{underlinded}.}
    \label{tab:asd_video}
\end{table}

Table \ref{tab:asd_video} shows the performance of \texttt{VITA} and \texttt{GenVIS} on video level on $\mathcal{D}^\text{val}_\text{ASD}$. Here, in contrast to image-level metrics, not one method clearly outperforms the other, however, since \texttt{GenVIS} still achieves competitive results, it remains our choice of VIS method on $\mathcal{D}_\text{ASD}$.

\end{document}